%% file: text_objseg_arxiv.tex
\documentclass[runningheads]{llncs}
\usepackage{graphicx}
\usepackage{tabularx}
\usepackage{amsmath,amssymb} 
\usepackage{color}
\usepackage[width=122mm,left=12mm,paperwidth=146mm,height=193mm,top=12mm,paperheight=217mm]{geometry}
\usepackage{xspace}
\newcolumntype{Y}{>{\centering\arraybackslash}X}

\begin{document}
\input{defines}
\input{macros}
\pagestyle{headings}
\mainmatter

\newcommand{\refexp}[1]{\emph{``#1''}}

\title{Segmentation from Natural Language Expressions} 

\titlerunning{Segmentation from Natural Language Expressions}

\authorrunning{Ronghang Hu, Marcus Rohrbach, Trevor Darrell}

\author{
Ronghang Hu$^{1}$\quad Marcus Rohrbach$^{1,2}$ \quad Trevor Darrell$^{1}$\\
\texttt{\{ronghang, rohrbach, trevor\}@eecs.berkeley.edu}
}

\institute{
 $^{1}$UC Berkeley EECS,  CA, United States\\
 $^{2}$ICSI, Berkeley, CA, United States\\
 }

\maketitle

\begin{abstract}
In this paper we approach the novel problem of segmenting an image based on a natural language expression. This is different from traditional semantic segmentation over a predefined set of semantic classes, as \eg, the phrase \refexp{two men sitting on the right bench} requires segmenting only the two people on the right bench and no one standing or sitting on another bench. Previous approaches suitable for this task were limited to a fixed set of categories and/or rectangular regions. To produce pixelwise segmentation for the language expression, we propose an end-to-end trainable recurrent and convolutional network model that jointly learns to process visual and linguistic information. In our model, a recurrent LSTM network is used to encode the referential expression into a vector representation, and a fully convolutional network is used to a extract a spatial feature map from the image and output a spatial response map for the target object. We demonstrate on a benchmark dataset that our model can produce quality segmentation output from the natural language expression, and outperforms baseline methods by a large margin.
\end{abstract}

\section{Introduction}

\begin{figure}[t]
\centering
\begin{tabularx}{\linewidth}{YYYY}
\includegraphics[width=0.24\textwidth]{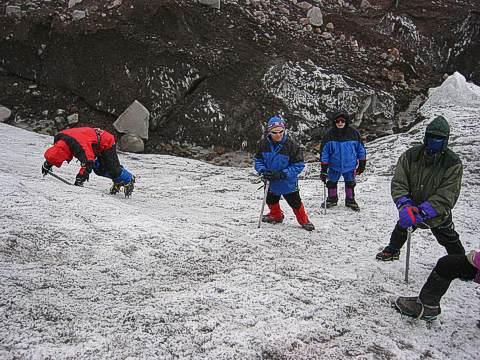} &
\includegraphics[width=0.24\textwidth]{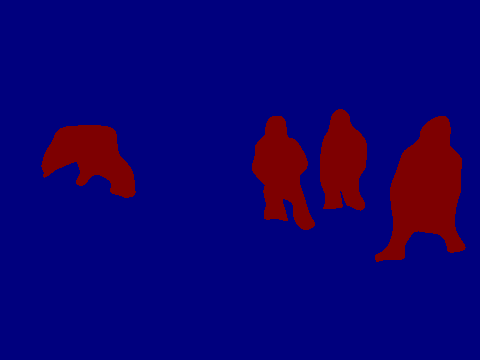} &
\includegraphics[width=0.24\textwidth]{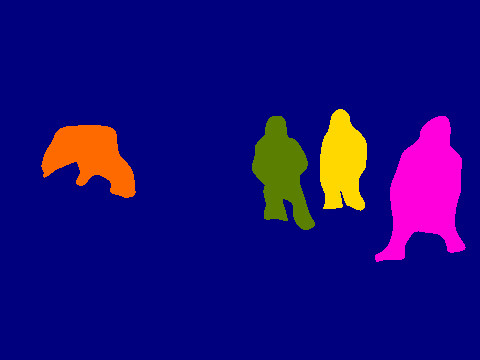} &
\includegraphics[width=0.24\textwidth]{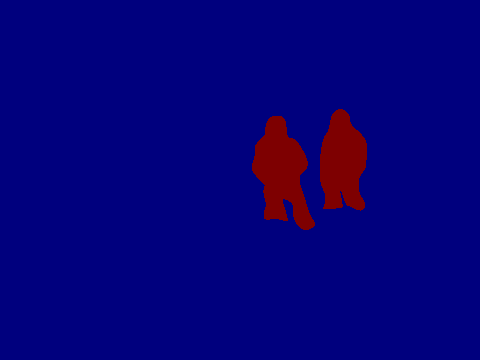} \\
(a) input image &
(b) object class segmentation of class \textbf{\textit{people}} &
(c) object instance segmentation of class \textbf{\textit{people}} &
(d) segmentation from expression \refexp{people in blue coat} \\
\end{tabularx}
\caption{In this work we approach the novel problem of \emph{segmentation from natural language expressions}, which is different from traditional semantic image segmentation and object instance segmentation, as visualized in this figure.}
\label{fig:teaser}
\end{figure}

Semantic image segmentation is a core problem in computer vision and significant progress has been made using large visual datasets and rich representations based on convolution neural networks \cite{long2015fully,carreira2012semantic,chen2014semantic,zheng2015conditional,noh2015learning,yu2015multi}. Although these existing segmentation methods can predict precise pixelwise masks for query categories like ``train'' or ``cat'', they are not capable of predicting segmentation for more complicated queries such as the natural language expression ``the two people on the right side of the car wearing black shirt''.

In this paper we address the following problem: given an image and a natural language expression that describes a certain part of the image, we want to segment the corresponding region(s) that covers the visual entities described by the expression. For example, as shown in \figref{fig:teaser} (d), for the phrase \eg \refexp{people in blue coat} we want to predict a segmentation that covers the two people in the middle wearing blue coat, but not the other two people. This problem is related to but different from the core computer vision problems of \emph{semantic segmentation} (\eg PASCAL VOC segmentation challenge on 20 object classes \cite{pascal-voc-2012}), which is concerned with predicting the pixelwise label for a predefined set of object or stuff categories (\figref{fig:teaser}, b), and \emph{instance segmentation} (\eg \cite{hariharan2014simultaneous}), which additionally distinguishes different instances of an object class (\figref{fig:teaser}, c). It also differs from language-independent foreground segmentation (\eg \cite{rother2004grabcut}), where the goal is to generate a mask over the foreground (or the most salient) object. Instead of assigning a semantic label to every pixel in the image as in semantic image segmentation, the goal in this paper is to produce a segmentation mask for the visual entities of interest based on the given expression. Rather than being fixed on a set of object and stuff categories, natural language descriptions may involve also attributes such as \refexp{black} and \refexp{smooth}, actions such as \refexp{running}, spatial relationships such as \refexp{on the right} and interactions between different visual entities such as \refexp{the person who is riding a horse}.

\begin{figure}[t]
\centering
\includegraphics[width=0.9\textwidth]{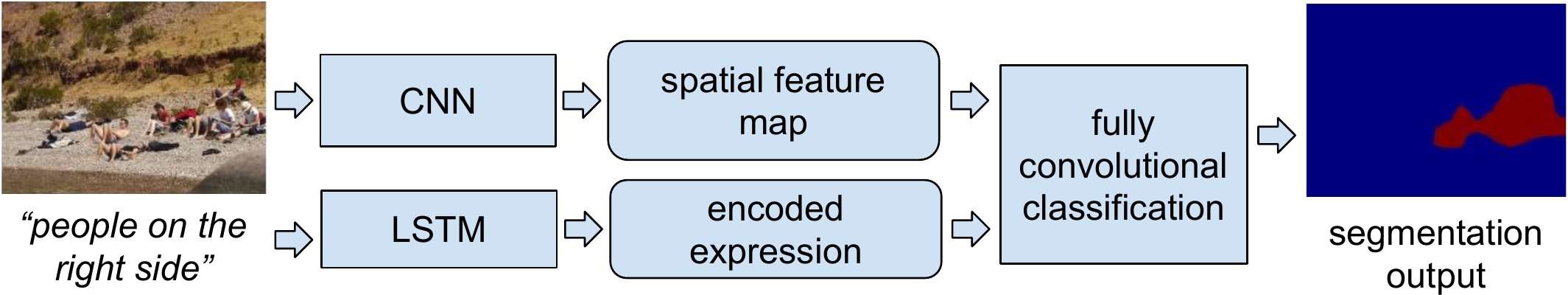}
\caption{Overview of our method for segmentation from natural language expressions.}
\label{fig:method_simple}
\end{figure}

The task of segmenting an image from natural language expressions has a wide range of applications, such as building language-based human-robot interface to give instructions like \refexp{pick up the jar on the table next to the apples} to a robot. Here, it is important to be able to use multi-word referential expressions to distinguish between different object instances but also important to get a precise segmentation in contrast to just a bounding box, especially for non-grid-aligned objects (see \eg Figure \ref{fig:method_simple}). This could also be interesting for interactive photo editing where one could refer with natural language to certain parts or objects of the image to be manipulated, \eg \refexp{blur the person with a red shirt}, or referring to parts of your meal to estimate their nutrition, \refexp{two large pieces of bacon}, to decide better if one should eat it rather than the full meal as in \cite{meyers15iccv}. 

As described in more details in Section \ref{sec:related_work}, prior methods suitable for this task were limited to resolving only a bounding box in the image \cite{hu2015natural,rohrbach2015grounding,mao2015generation}, and/or were limited to a fixed set of categories determined {\em a priori} \cite{long2015fully,chen2014semantic,zheng2015conditional,yu2015multi}. In this paper, we propose an end-to-end trainable recurrent convolutional network model that jointly learns to process visual and linguistic information, and produces segmentation output for the target image region described by the natural language expression, as illustrated in Figure \ref{fig:method_simple}.
We encode the expression into a fixed-length vector representation through a recurrent LSTM network, and use a convolutional network to extract a spatial feature map from the image. The encoded expression and the feature map are then processed by a multi-layer classifier network in a fully convolutional manner to produce a coarse response map, which is upsampled with deconvolution \cite{long2015fully,noh2015learning} to obtain a pixel-level segmentation mask of the target image region.
Experimental results on a benchmark dataset demonstrate that our model can generate quality segmentation predictions from natural language expressions, and outperforms baseline methods significantly. Our model is trained using standard back-propagation, and is much more efficient at test time than previous approaches relying on scoring each bounding box.

\section{Related work}\label{sec:related_work}

Our work is related to several areas as follows.

\textbf{Localizing objects with natural language.}
Our work is related to recent work on object localization with natural language, where the task is to localize a target object in a scene from its natural language description (by drawing a bounding box over it). The methods reported in \cite{hu2015natural} and \cite{mao2015generation} build upon image captioning frameworks such as LRCN \cite{donahue2015long} or mRNN \cite{mao2014explain}, and localize objects by selecting the bounding box where the expression has the highest probability. Our model differs from \cite{hu2015natural} and \cite{mao2015generation} in that we do not have to learn to generate expressions from image regions. In \cite{rohrbach2015grounding}, the authors propose a model to localize a textual phrase by attending to a region on which the phrase can be best reconstructed. In \cite{plummer15iccv}, Canonical Correlation Analysis (CCA) is used to learn a joint embedding space of visual features and words, and given a natural language query, the corresponding target object is localized by finding the closest region to the text sequence in the joint embedding space. 

To the best of our knowledge, all these previous localization methods can only return a bounding box of the target object, and no prior work has learned to directly output a segmentation mask of an object given a natural language description as query. As a comparison, in Section \ref{sec:baselines} we also evaluate using foreground segmentation over the bounding box prediction from \cite{hu2015natural} and \cite{rohrbach2015grounding}.

\textbf{Fully convolutional network for segmentation.} Fully convolutional networks are convolutional neural networks consisting of only convolutional (and pooling) layers, which are the state-of-the-art method for semantic segmentation over a pre-defined set of semantic categories \cite{long2015fully,chen2014semantic,zheng2015conditional,yu2015multi}. A nice property of fully convolutional networks is that spatial information is preserved in the output, which makes these networks suitable for segmentation tasks that require spatial grid output. In our model, both feature extraction and segmentation output are performed through fully convolutional networks. We also use a fully convolution network for per-word segmentation as a baseline in Section \ref{sec:baselines}.

\textbf{Attention and visual question answering.} Recently, attention models have been used in several areas including image recognition, image captioning and visual question answering. In \cite{xu2015show}, image captions are generated through focusing on a specific image region for each word. In recent visual question answering models \cite{yang2015stacked,xu2015ask}, the answer is determined through attending to one or multiple image regions. The authors of \cite{andreas2016learning} propose a visual question answering method that can learn to answer object reference questions like \refexp{where is the black cat} through parsing the sentence and generating attention maps for \refexp{black} and ``cat''.

These attention models are related to our work as they also learn to generate spatial grid ``attention maps'' which often cover the objects of interest. However, these attention models differ from our work as they only learn to generate coarse spatial outputs and the purpose of the attention map is to facilitate other tasks such as image captioning, rather than precisely segment out the object.

\section{Our model}

\begin{figure}[t]
\centering
\includegraphics[width=0.9\textwidth]{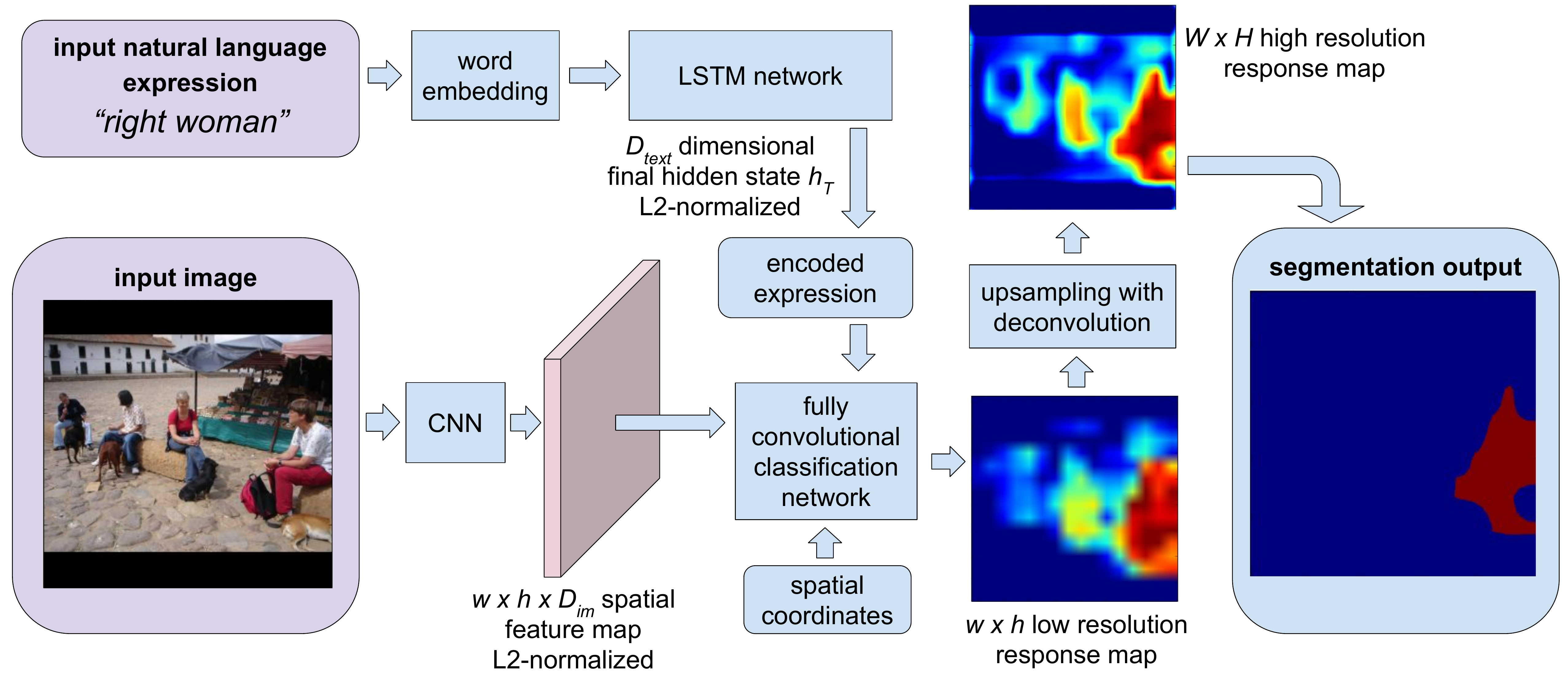}
\caption{Our model for segmentation from natural language expressions consists of three main components: an expression encoder based upon a recurrent LSTM network, a fully convolutional network to generate a spatial feature map, and a fully convolutional classification and upsampling network to predict pixelwise segmentation.}
\label{fig:method_overview}
\end{figure}

Given an image and a natural language expression as query, the goal is to output a segmentation mask for the visual entities described by the expression. This problem requires both visual and linguistic understanding of the image and the expression. To accomplish this goal, we propose a model with three main components: a natural language expression encoder based on a recurrent LSTM network, a fully convolutional network to extract local image descriptors and generate a spatial feature map, and a fully convolutional classification and upsampling network that takes as input the encoded expression and the spatial feature map and outputs a pixelwise segmentation mask. Figure \ref{fig:method_overview} shows the outline of our method; we introduce the details of these components in Section \ref{sec:encode_image}, \ref{sec:encode_text} and \ref{sec:cls_upsample}. The network architecture for feature map extraction and classification is similar to the FCN model \cite{long2015fully}, which has been shown effective for semantic image segmentation. 

Compared with related work \cite{hu2015natural,mao2015generation}, we do not explicitly produce a word sequence corresponding to object descriptions given a visual representation, since we are interested in predicting image segmentation from an expression rather than predicting the expression. In this way, our model has less parameters compared with \cite{hu2015natural,mao2015generation} as it does not have to learn to predict the next word, which can be a hard task.

\subsection{Spatial feature map extraction}\label{sec:encode_image}

Given an image of a scene, we want to obtain a discriminative feature representation of it while preserving the spatial information in the representation so that it is easier to predict a spatial segmentation mask. This is accomplished through a fully convolutional network model similar to FCN-32s \cite{long2015fully}, where the image is fed through a series of convolutional (and pooling) layers to obtain a spatial map output as feature representation. Given an input image of size $W \times H$, we use a convolutional network on the image to obtain a $w \times h$ spatial feature map, with each position on the feature map containing $D_{im}$ channels ($D_{im}$ dimensional local descriptors).

For each spatial location on the feature map, we apply L2-normalization to the $D_{im}$ dimensional local descriptor at that position in order to obtain a more robust feature representation. In this way, we can extract a $w \times h \times D_{im}$ spatial feature map as the representation for each image.

Also, to allow the model to reason about spatial relationships such as ``right woman'' in Figure \ref{fig:method_overview}, two extra channels are added to the feature maps: the $x$ and $y$ coordinate of each spatial location. We use relative coordinates, where the upper left corner and the lower right corner of the feature map are represented as $(-1, -1)$ and $(+1, +1)$, respectively. In this way, we obtain a $w \times h \times (D_{im}+2)$ representation containing local image descriptors and spatial coordinates.

In our implementation, we adopt the VGG-16 architecture \cite{simonyan2014very} as our fully convolutional network by treating fc6, fc7 and fc8 as convolutional layers, which outputs $D_{im}=1000$ dimensional local descriptors. The resulting feature map size is $w= W / s$ and $h= H / s$, where $s = 32$ is the pixel stride on fc8 layer output. The units on the spatial feature map have a very large receptive field of 384 pixels, so our method has the potential to aggregate contextual information from nearby regions, which can help to reason about interaction between visual entities, such as ``the man next to the table''.

\subsection{Encoding expressions with LSTM network}\label{sec:encode_text}

For the input natural language expression that describes an image region, we would like to represent the text sequence as a vector since it is easier to process fixed-length vectors than variable-length sequences. To achieve this goal, we take the encoder approach in sequence to sequence learning methods \cite{sutskever2014sequence,cho2014properties}. In our encoder for the natural language expression, we first embed each word into a vector through a word embedding matrix, and then use a recurrent Long-Short Term Memory (LSTM) \cite{hochreiter1997long} network with $D_{text}$ dimensional hidden state to scan through the embedded word sequence. For a text sequence $S=(w_1, ..., w_T)$ with $T$ words (where $w_t$ is the vector embedding for the $t$-th word), at each time step $t$, the LSTM network takes as input the embedded word vector $w_t$ from the word embedding matrix. At the final time step $t=T$ after the LSTM network have seen the whole text sequence, we use the hidden state $h_T$ in LSTM network as the encoded vector representation of the expression. Similar to Section \ref{sec:encode_image}, we also L2-normalize the $D_{text}$ dimensions in $h_T$. We use a LSTM network with $D_{text}=1000$ dimensional hidden state in our implementation.

\subsection{Spatial classification and upsampling}\label{sec:cls_upsample}

After extracting the spatial feature map from the image in Section \ref{sec:encode_image} and the encoded expression $h_T$ in Section \ref{sec:encode_text}, we want to determine whether or not each spatial location on the feature map belongs the foreground (the visual entities described by the natural language expression). In our model, this is done by a fully convolutional classifier over the local image descriptor and the encoded expression.
We first tile and concatenate $h_T$ to the local descriptor at each spatial location in the spatial grid to obtain a $w \times h \times D^*$ (where $D^*=D_{im}+D_{text}+2$) spatial map containing both visual and linguistic features. Then, we train a two-layer classification network, with a $D_{cls}$ dimensional hidden layer, which takes at input the $D^*$ dimensional representation and output a score to indicate whether a spatial location belong to the target image region or not. We use $D_{cls}=500$ in our implementation.

This classification network is applied in a fully convolutional way over the underlying $w \times h$ feature map as two $1 \times 1$ convolutional layers (with ReLU none linearity between them). The fully convolutional classification network outputs a $w \times h$ coarse \textit{low-resolution response map} containing classification scores, which can be seen as a low-resolution segmentation of the referential expression, as shown in Figure \ref{fig:method_overview}.

In order obtain a segmentation mask with higher resolution, we further perform upsampling through deconvolution (swapping the forward and backward pass of convolution operation) \cite{long2015fully,noh2015learning}. Here we use a $2s \times 2s$ deconvolution filter with stride $s$ (where $s = 32$ for the VGG-16 network architecture we use), which is similar to the FCN-32s model \cite{long2015fully}. The deconvolution operation produces a $W \times H$ \textit{high resolution response map} that has the same size as the input image, and the values on the high resolution response map represent the confidence of whether a pixel belongs to the target object. We use the pixelwise classification results (\ie whether or not a value on the response map is greater than 0) as the final segmentation prediction.

At training time, each training instance in our training set is a tuple $(I, S, M)$, where $I$ is an image, $S$ is a natural language expression describing a region within that image, and $M$ is a binary segmentation mask of that region. The loss function during training is defined as the average over pixelwise loss
\begin{equation}
Loss = \frac{1}{W H}\sum_{i=1}^W \sum_{j=1}^H  L(v_{ij}, M_{ij})
\end{equation}
where $W$ and $H$ are image width and height, $v_{ij}$ is the response value (score) on the high resolution response map and $M_{ij}$ is the binary ground-truth label at pixel $(i,j)$. $L$ is the per-pixel weighed logistic regression loss as follows
\begin{equation}
L(v_{ij}, M_{ij}) = \begin{cases}
\alpha_f \log(1 + \exp(-v_{ij})) & \text{if } M_{ij} = 1 \\
\alpha_b \log(1 + \exp( v_{ij})) & \text{if } M_{ij} = 0 \\
\end{cases}
\end{equation}
where $\alpha_f$ and $\alpha_b$ are loss weights for foreground and background pixels. In practice, we find that training converges faster using higher loss weights for foreground pixels, and we use $\alpha_f = 3$ and $\alpha_b = 1$ in $L(v_{ij}, M_{ij})$.

The parameters in feature map extraction network are initialized from a VGG-16 network \cite{simonyan2014very} pretrained on the 1000-class ILSVRC classification task \cite{russakovsky2015imagenet}, the deconvolution filter for upsampling is initialized from bilinear interpolation. All other parameters in our model, including the word embedding matrix, the LSTM parameters and the classifier parameters, are randomly initialized. The whole network is trained with standard back-propagation using SGD with momentum.

\section{Experiments}\label{sec:exp-referit}

Compared with the widely used datasets in image segmentation such as PASCAL VOC \cite{pascal-voc-2012}, there are only a few publicly available datasets with natural language annotations over segmented image regions. In our experiments, we train and test our method on the ReferIt dataset \cite{kazemzadeh2014referitgame} with natural language descriptions of visual entities and their segmentation masks. The ReferIt dataset \cite{kazemzadeh2014referitgame} is built upon the IAPR TC-12 dataset \cite{grubinger2006iapr} and has 20,000 images. There are 130,525 expressions annotated on 96,654 segmented image regions (some regions are annotated with multiple expressions). In this dataset, the ground-truth segmentation comes from the SAIAPR-12 dataset \cite{escalante2010segmented}. The expressions in the ReferIt dataset are discriminative for the regions, as they were collected in a two-player game whose goal was to make the target region easily distinguishable through the expression from the rest of the image. At the time of writing, the ReferIt dataset \cite{kazemzadeh2014referitgame} is the biggest publicly available dataset that contains natural language expressions annotated on segmented image regions.

On this dataset, we use the same trainval and test split as in \cite{hu2015natural,rohrbach2015grounding}. There are 10,000 images for training and validation, and 10,000 images for testing. The annotated regions in the ReferIt dataset contains both ``object'' regions such as car, person and bottle and ``stuff'' regions such as sky, river and mountain.

Although \cite{mao2015generation} also collected a separate Google-RefExp dataset containing natural language expressions with segmented regions available from MS COCO dataset annotations \cite{lin2014microsoft}, this dataset only contains object annotations from 80 object categories in COCO, and does not have ``stuff'' regions such as snow. \footnote{At the time of this writing, the test split of Google-RefExp dataset has not been released. Evaluation on this dataset is a part of our future work.}

Since there has not been prior work that directly learns to predict segmentation based on natural language expressions as far as we know, to evaluate our method, we construct several strong baseline methods as described in Section \ref{sec:baselines}, and compare our approach with these methods.

\subsection{Baseline methods}\label{sec:baselines}

\textbf{Combination of per-word segmentation.}
In this baseline method, instead of first encoding the whole expression with a recurrent LSTM network, each word in the expression is segmented individually, and the per-word segmentation results are then combined to obtain the final prediction. This method can be seen as using a ``bag-of-word'' representation of the expression. We take the $N$ most frequently appearing words in ReferIt dataset (after manually removing some stop words like ``the'' and ``towards''), and train a FCN model \cite{long2015fully} to segment each word. Similar to the PASCAL VOC segmentation challenge \cite{pascal-voc-2012}, in this method, each word is treated as an independent semantic category. However, unlike in PASCAL VOC segmentation, here a pixel can belong to multiple categories (words) simultaneously and thus have multiple labels. During training, we generate a per-word pixelwise label map for each training sample (an image and an expression) in the training set. For a given expression, the corresponding foreground pixels are labeled with a $N$-dimensional binary vector $l$, where $l_i$ = 1 if and only if word $i$ is present in the expression, and background pixels are labeled with $l$ equal to all zeros. In our experiments, we use $N=500$ and initialize the network from a FCN-32s network pretrained on PASCAL VOC 2011 segmentation task \cite{long2015fully}, and train the whole network with a multi-label logistic regression loss over the words.

At test time, given an image and a natural language expression as input, the network outputs pixelwise score maps for the $N$ words, and the per-word scores are further combined to obtain the segmentation for the input expression. In our implementation, we experiment with three different approaches to combine the per-word segmentation: for those words (among the $N$-word list) that appear in the expression, we a) take the average of their scores or b) take the intersection of their prediction or c) take the union of their prediction. In some rare cases (2.83\% of the test samples), none of the words in the expression are among the $N$ most frequent words, and we do not output any segmentation for this expression, \ie all pixels are predicted as background.

\textbf{Foreground segmentation from bounding boxes.}
In this baseline method, we first use a localization method based on natural language input \cite{hu2015natural,rohrbach2015grounding} to obtain a bounding box localization of the given expression, and then extract the foreground segmentation from the bounding box using GrabCut \cite{rother2004grabcut}. Given an image and a natural language expression, we use two recently proposed methods SCRC \cite{hu2015natural} and GroundeR \cite{rohrbach2015grounding} to obtain a bounding box prediction from the image and the expression. In SCRC \cite{hu2015natural}, the authors use a model adapted from image captioning and localize a referential expression by finding the candidate bounding box where the expression receives the highest probability. In GroundeR \cite{rohrbach2015grounding}, an attention model over candidate bounding boxes is used to ground (localize) a referential expression, either in an unsupervised manner by finding the region that can best reconstruct the expression, or in a supervised manner to directly train the model to attend to the best bounding box. Following \cite{hu2015natural,rohrbach2015grounding}, we use 100 top-scoring EdgeBox \cite{zitnick2014edge} proposals as a set of candidate bounding boxes for each image. At test time, given an input expression, we compute the scores of the 100 EdgeBox proposals using SCRC \cite{hu2015natural} or GroundeR \cite{rohrbach2015grounding}, and evaluate two approaches: either using the entire rectangular region of the highest scoring bounding box, or the foreground segmentation from it using GrabCut \cite{rother2004grabcut}. We use the supervised version of \cite{rohrbach2015grounding} in our experiments.

\textbf{Classification over segmentation proposals.}
In this baseline method, we first extract a set of candidate segmentation proposals using MCG \cite{arbelaez2014multiscale}, and then train a binary classifier to determine whether or not a candidate segmentation proposal matches the expression. We use a similar pipeline in this baseline as in the supervised version of \cite{rohrbach2015grounding}. First, visual features are extracted from each proposal and concatenated with the encoded sentence. Then, a classification network is trained on concatenated features to classify a segmentation proposal into foreground or background. We use 100 top-scoring segmentation proposals from MCG, and extract visual features from each segmentation by first resizing it to $224\times 224$ (those pixels outside the segmentation region are filled with channel mean) and then extracting visual feature from the resized segmentation using a VGG-16 network pretrained on ILSVRC classification task. The whole network is then trained end-to-end. The main difference between this baseline and our method is that our method performs pixelwise classification through a fully convolutional network, while this baseline requires another proposal method to obtain candidate regions.

\textbf{Whole image.} As an additional trivial baseline, we also evaluate using the whole image as a segmentation for every expression.

\subsection{Evaluation on ReferIt dataset}

\begin{table}[t]
\begin{center}
\begin{tabular}{|l|r|r|r|r|r||r|}
\hline
\textbf{Method} & \textit{prec}@0.5 & \textit{prec}@0.6 & \textit{prec}@0.7 & \textit{prec}@0.8 & \textit{prec}@0.9 & overall IoU \\
\hline
whole image    										&  5.07\% & 2.85\% & 1.58\% & 0.81\% & 0.41\% & 15.12\% \\
per-word average    								    & 10.97\% & 5.94\% & 2.35\% & 0.45\% & 0.00\% & 27.23\% \\
per-word intersection 								&  9.58\% & 5.35\% & 2.20\% & 0.43\% & 0.00\% & 26.69\% \\
per-word union     									& 10.46\% & 5.65\% & 2.28\% & 0.44\% & 0.00\% & 19.37\% \\
SCRC \cite{hu2015natural} bbox 				        &  9.73\% & 4.43\% & 1.51\% & 0.27\% & 0.03\% & 21.72\% \\
SCRC \cite{hu2015natural} grabcut 				    & 11.91\% & 7.71\% & 4.33\% & 1.78\% & 0.36\% & 17.84\% \\
GroundeR \cite{rohrbach2015grounding} bbox 			& 11.08\% & 6.20\% & 2.74\% & 0.78\% & 0.20\% & 20.50\% \\
GroundeR \cite{rohrbach2015grounding} grabcut 		& 14.09\% & 9.62\% & 5.78\% & 2.65\% & 0.62\% & 20.09\% \\
MCG classification    								& 12.72\% & 9.88\% & 7.38\% & 4.73\% & 1.88\% & 18.08\% \\
\hline
Ours (low resolution)      & 29.54\% & 21.61\% & 13.69\% & 5.94\% & 0.75\% & 45.57\%  \\
Ours (high resolution)       & \textbf{34.02\%} & \textbf{26.71\%} & \textbf{19.32\%} & \textbf{11.63\%} &  \textbf{3.92\%} & \textbf{48.03\%}  \\
\hline
\end{tabular}
\end{center}
\caption{The performance of our model and baseline methods on the ReferIt dataset under precision metric and overall IoU metric. See Section \ref{sec:exp-referit} for details.}
\label{tab:results_referit}
\end{table}

We train our model and the baseline methods in Section \ref{sec:baselines} on the 10,000 trainval images in the ReferIt dataset \cite{kazemzadeh2014referitgame} (leaving out a small proportion for validation), following the same split as in \cite{hu2015natural}. In our implementation, we resize and pad all images and ground-truth segmentation to a fixed size $W \times H$ (where we set $W = H = 512$), keeping their aspect ratio and padding the outside regions with zero, and map the segmentation output back to the original image size to obtain the final segmentation.

In our experiments, we use a two-stage training strategy: we first train a low resolution version of our model, and then fine-tune from it to obtain the final high resolution model (\ie our full model in Figure \ref{fig:method_overview}). In our low resolution version, we do not add the deconvolution filter in Section \ref{sec:cls_upsample}, so the model only outputs a $w \times h = 16 \times 16$ coarse response map in Figure \ref{fig:method_overview}. We also downsample the ground-truth label to $w \times h$ and directly train on the coarse response map to match the downsampled label. After training the low resolution model, we construct our final high resolution model by adding a $2s \times 2s$ deconvolution filter with stride $s=32$, as described in Section \ref{sec:cls_upsample}, and initialize the filter weights from bilinear interpolation (all other parameters are initialized from low resolution model). The high resolution model is then fine-tuned on the training set using $W \times H$ ground-truth segmentation mask labels. We empirically find this two stage training converges faster than directly training our full model to predict $W \times H$ high resolution segmentation.

We evaluate the performance of our model and the baseline methods in Section \ref{sec:baselines} on the 10,000 images in the test set. The following two metrics are used for evaluation: the \textit{overall intersection-over-union} (overall IoU) metric and the \textit{precision} metric. The overall IoU is the total intersection area divided by the total union area, where both intersection area and union area are accumulated over all test samples (each test sample is an image and a referential expression). Although the overall IoU metric is the standard metric used in PASCAL VOC segmentation \cite{hu2015natural}, our evaluation is slighly different as we would like to measure how accurate the model can segment the foreground region described by the input expression against the background, and the overall IoU metric favors large regions like sky and ground. So we also evaluate with the precision metric at 5 different IoU thresholds from easy to hard: 0.5, 0.6, 0.7, 0.8, 0.9. The precision metric is the percentage of test samples where the IoU between prediction and ground-truth passes the threshold. For example, precision@0.5 is the percentage of expressions where the predicted segmentation overlaps with the ground-truth region by at least 50\% IoU.

\begin{figure}[t]
\centering
\begin{tabularx}{0.95\linewidth}{YYYY}
input image & our model & per-word & GroundeR \cite{rohrbach2015grounding} \\ \hline
\end{tabularx} \\
\small{query expression=\refexp{person}} \\
\includegraphics[trim = 56mm 38mm 45mm 35mm, clip=true,width=0.90\textwidth,height=\textheight,keepaspectratio]{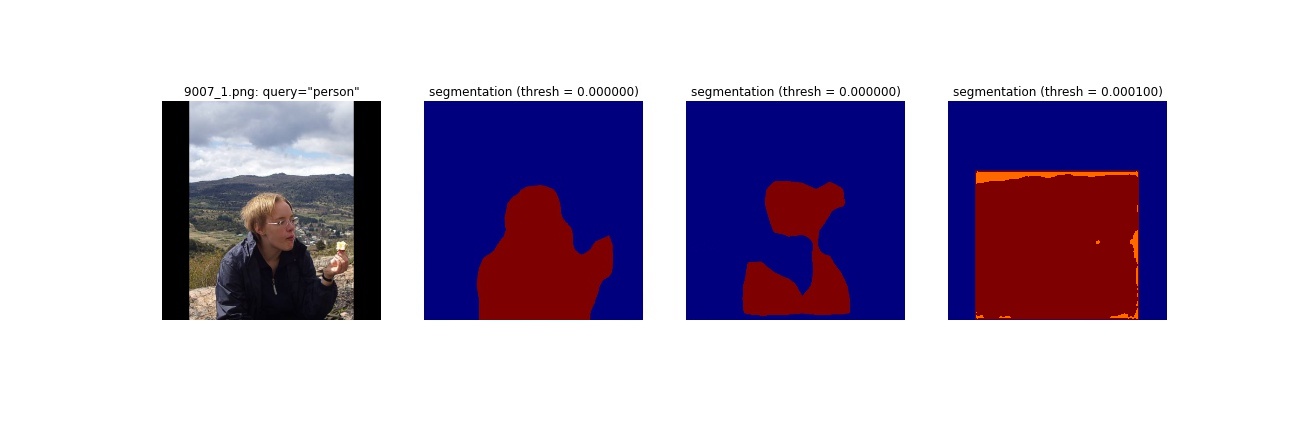} \\
\small{query expression=\refexp{car}} \\
\includegraphics[trim = 56mm 53mm 45mm 48mm, clip=true,width=0.90\textwidth,height=\textheight,keepaspectratio]{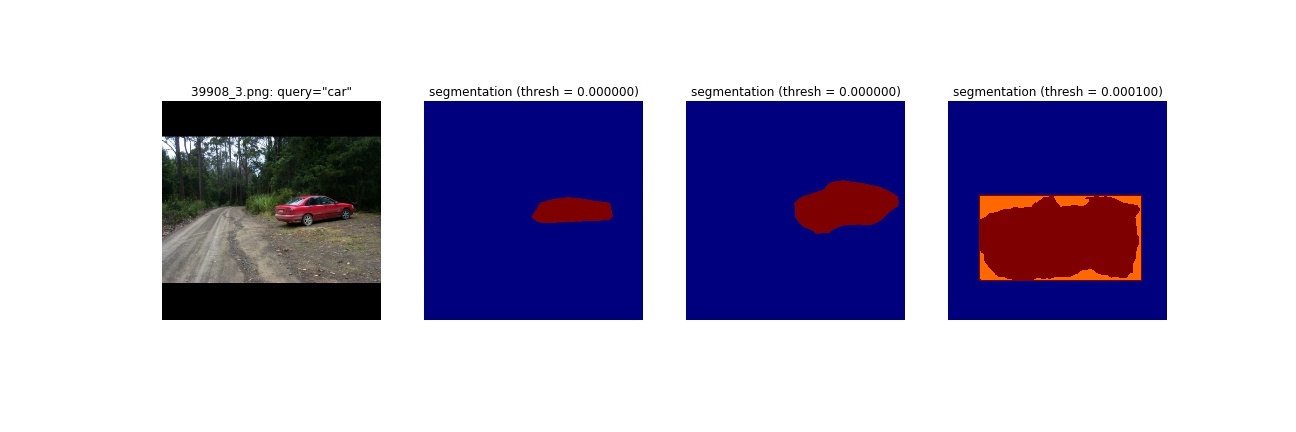} \\
\small{query expression=\refexp{the water at the bottom of the picture}} \\
\includegraphics[trim = 56mm 50mm 45mm 45mm, clip=true,width=0.90\textwidth,height=\textheight,keepaspectratio]{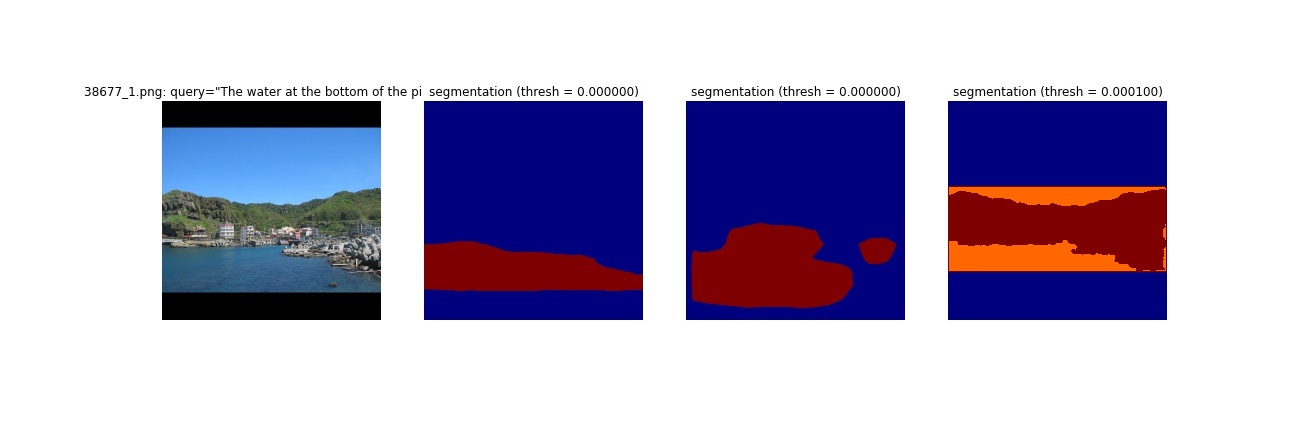} \\
\small{query expression=\refexp{people}} \\
\includegraphics[trim = 56mm 50mm 45mm 45mm, clip=true,width=0.90\textwidth,height=\textheight,keepaspectratio]{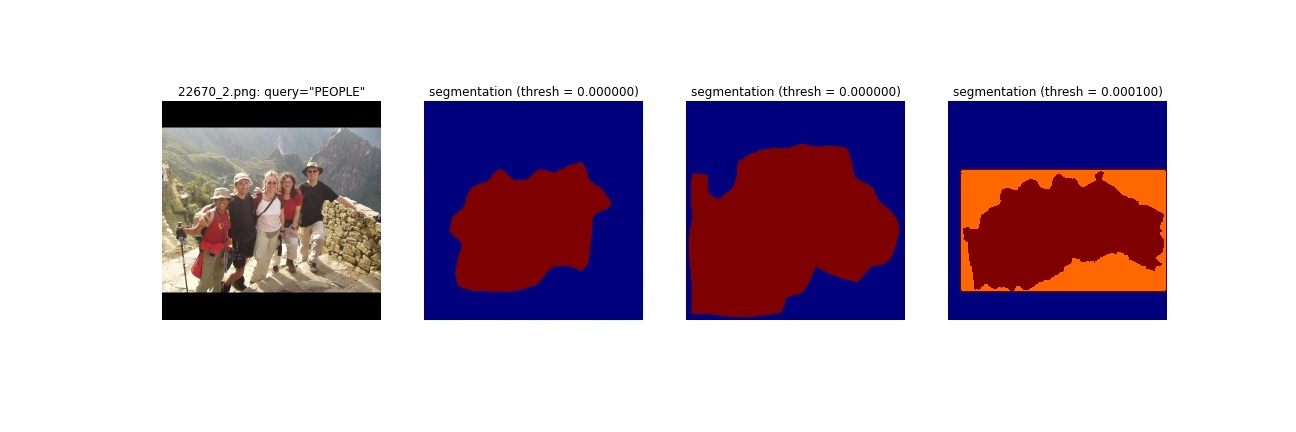} \\
\small{query expression=\refexp{person on raft}} \\
\includegraphics[trim = 56mm 53mm 45mm 48mm, clip=true,width=0.90\textwidth,height=\textheight,keepaspectratio]{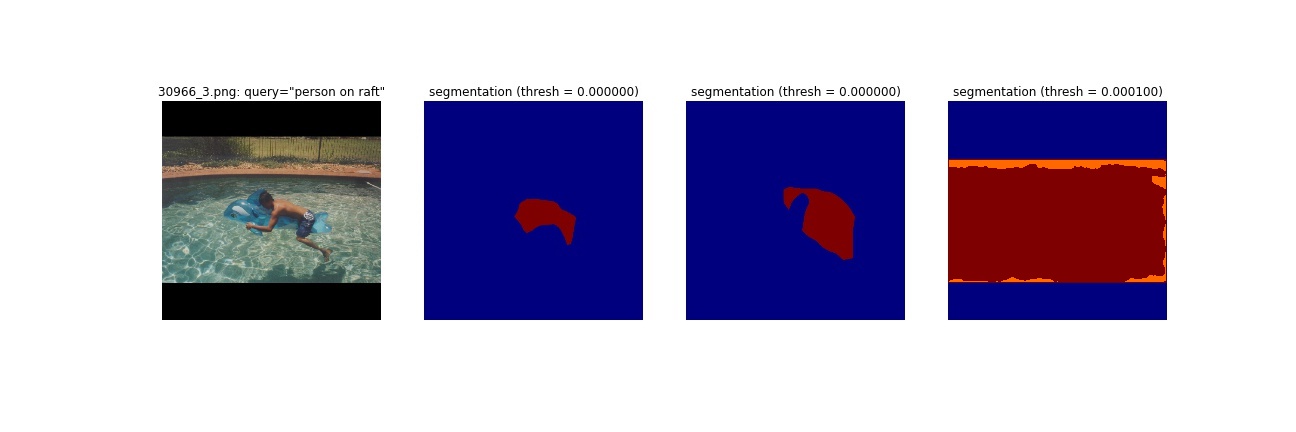} \\
\caption{Segmentation examples using our model and baseline methods. For GroundeR \cite{rohrbach2015grounding}, the bounding box prediction is in orange and GrabCut segmentation is in red.}
\label{fig:comparison}
\end{figure}

\textbf{Results.}
The main results for our evaluation are summarized in Table \ref{tab:results_referit}. By simply returning the whole image, one already gets 15\% overall IoU. This is partially due to the fact that the ReferIt dataset contains some large regions such as ``sky'' and ``city'' and the overall IoU metric put more weights on large regions. However, as expected, the whole image baseline has the lowest precision.

It can be seen from Table \ref{tab:results_referit} that one can get a reasonable overall IoU through per-word segmentation and combining the results from each word. Among the three different ways to combine the per-word results in Section \ref{sec:baselines}, it works best to average the scores from each word. Using the whole bounding box prediction from SCRC \cite{hu2015natural} (``SCRC bbox'') or GroundeR \cite{rohrbach2015grounding} (``GroundeR bbox'') achieves comparable precision to averaging per-word segmentation, while they are worse in terms of overall IoU, and using classification over segmentation proposals from MCG (``MCG classification'') leads to slightly higher precision than these two methods. Also, it can be seen that using GrabCut \cite{rother2004grabcut} to segment the foreground from bounding boxes (``SCRC grabcut'' and ``GroundeR grabcut'') results in higher precision for both SCRC and GroundeR than using the entire bounding box region. We believe that the precision metric is more reflective for the performance of segmentation methods over natural language expressions, since in real applications, one would often care more about how often a referential expression is correctly segmented.

Our model outperforms all the baseline methods by a large margin under both precision metric and overall IoU metric. In Table \ref{tab:results_referit}, the second last row (``low resolution'') corresponds to directly using bilinear upsampling over the coarse response map from our low resolution model, and the last row (``high resolution'') shows the performance of our full model. It can be seen that our final model achieves significantly higher precision and overall IoU, compared with the baseline methods. Figure \ref{fig:comparison} shows some segmentation examples using our model and baseline methods.

\begin{figure}
\centering
\begin{tabularx}{0.95\linewidth}{YYYY}
input image & response map & our prediction & ground-truth \\ \hline
\end{tabularx} \\
\small{query expression=\refexp{bird on the left}} \\
\includegraphics[trim = 56mm 53mm 45mm 48mm, clip=true,width=0.90\textwidth,height=\textheight,keepaspectratio]{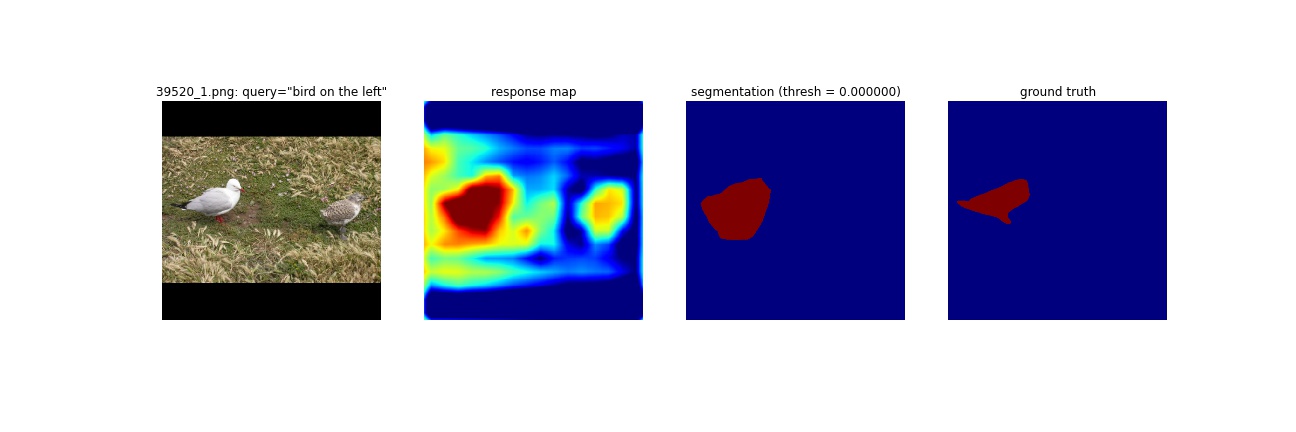} \\
\small{query expression=\refexp{three people on right}} \\
\includegraphics[trim = 56mm 50mm 45mm 45mm, clip=true,width=0.90\textwidth,height=\textheight,keepaspectratio]{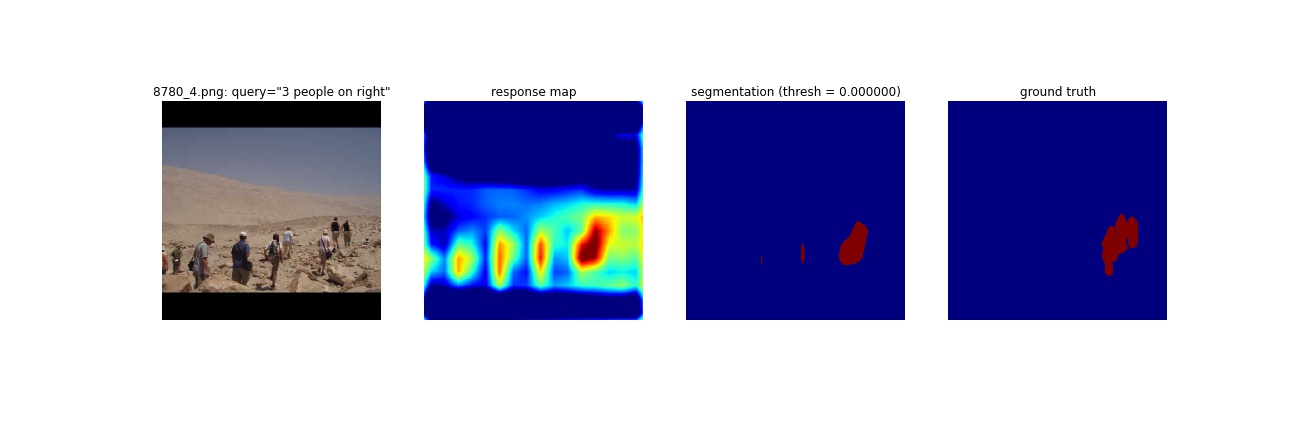} \\
\small{query expression=\refexp{anyone}} \\
\includegraphics[trim = 56mm 38mm 45mm 35mm, clip=true,width=0.90\textwidth,height=\textheight,keepaspectratio]{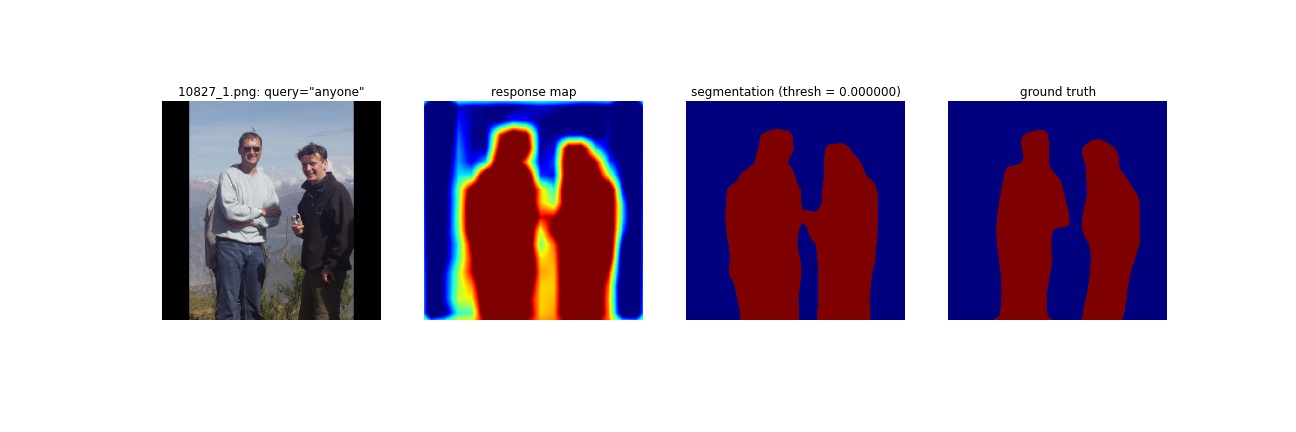} \\
\small{query expression=\refexp{big black suitcase bottom left}} \\
\includegraphics[trim = 56mm 50mm 45mm 45mm, clip=true,width=0.90\textwidth,height=\textheight,keepaspectratio]{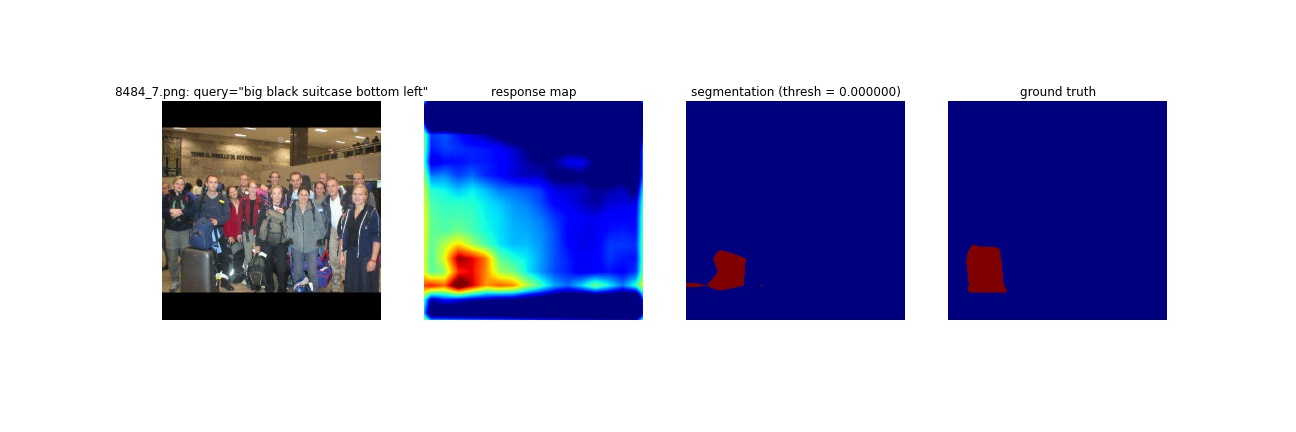} \\
\small{query expression=\refexp{man far right}} \\
\includegraphics[trim = 56mm 50mm 45mm 45mm, clip=true,width=0.90\textwidth,height=\textheight,keepaspectratio]{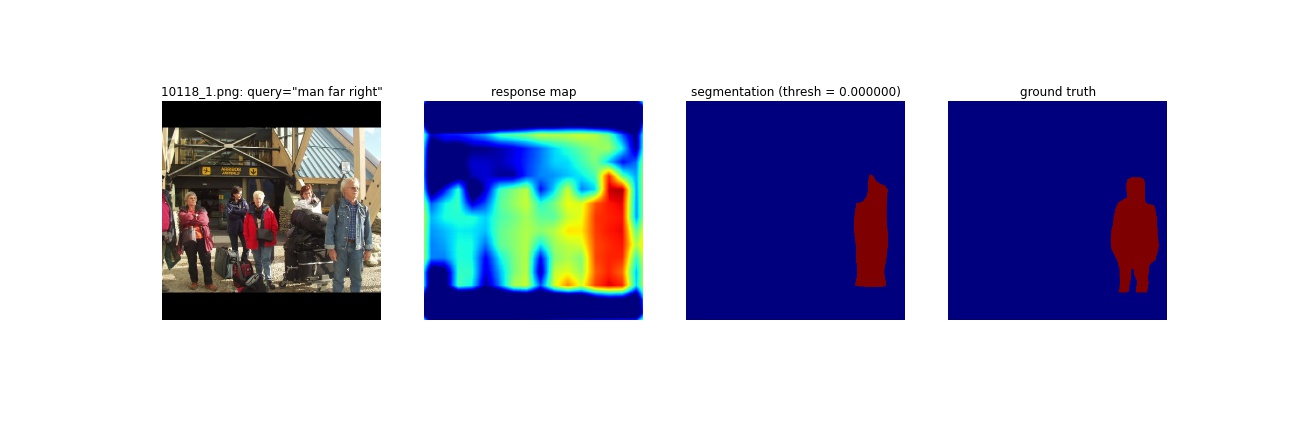} \\
\small{query expression=\refexp{bike}} \\
\includegraphics[trim = 56mm 53mm 45mm 49mm, clip=true,width=0.90\textwidth,height=\textheight,keepaspectratio]{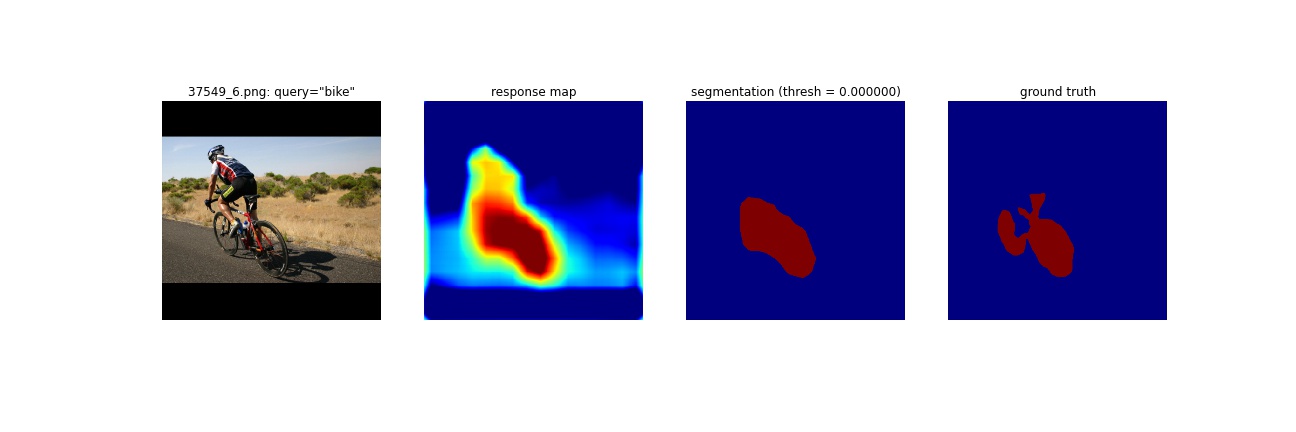} \\
\small{query expression=\refexp{guy in front}} \\
\includegraphics[trim = 56mm 53mm 45mm 49mm, clip=true,width=0.90\textwidth,height=\textheight,keepaspectratio]{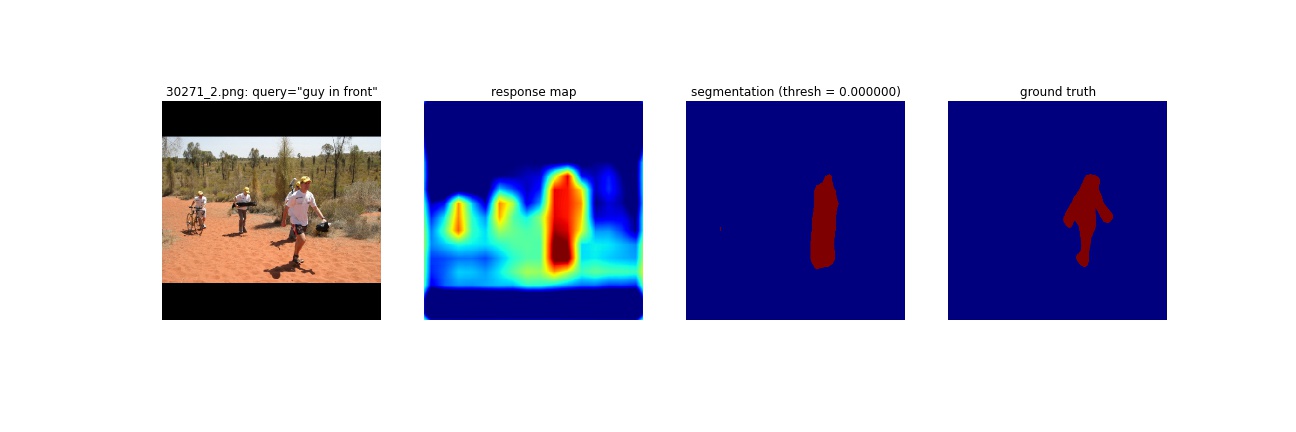} \\
\small{query expression=\refexp{left cactus}} \\
\includegraphics[trim = 56mm 50mm 45mm 45mm, clip=true,width=0.90\textwidth,height=\textheight,keepaspectratio]{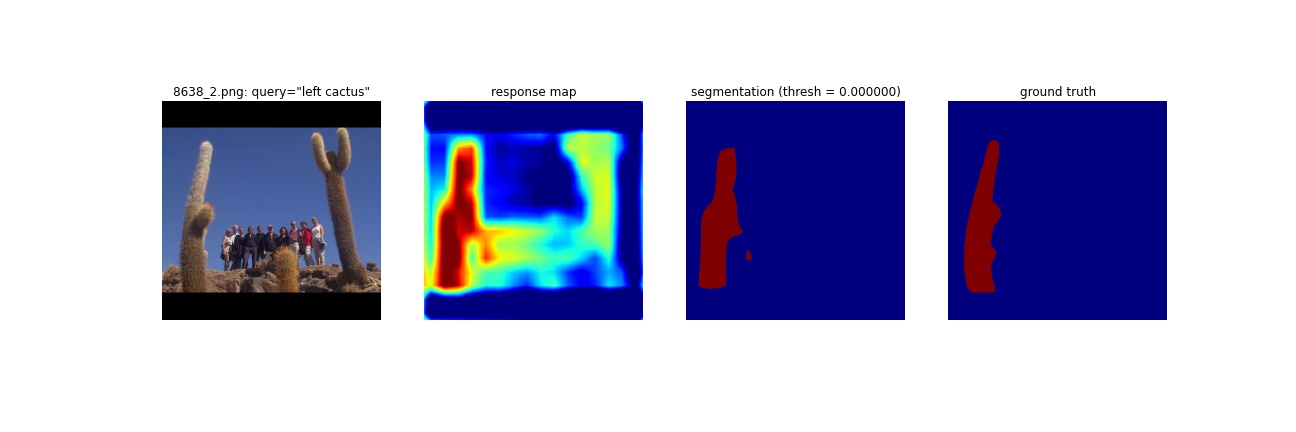} \\
\caption{Segmentation examples on object regions in the ReferIt dataset.}
\label{fig:sample_object}
\end{figure}

\begin{figure}
\centering
\begin{tabularx}{0.95\linewidth}{YYYY}
input image & response map & our prediction & ground-truth \\ \hline
\end{tabularx} \\
\small{query expression=\refexp{sky above the bridge}} \\
\includegraphics[trim = 56mm 50mm 45mm 45mm, clip=true,width=0.90\textwidth,height=\textheight,keepaspectratio]{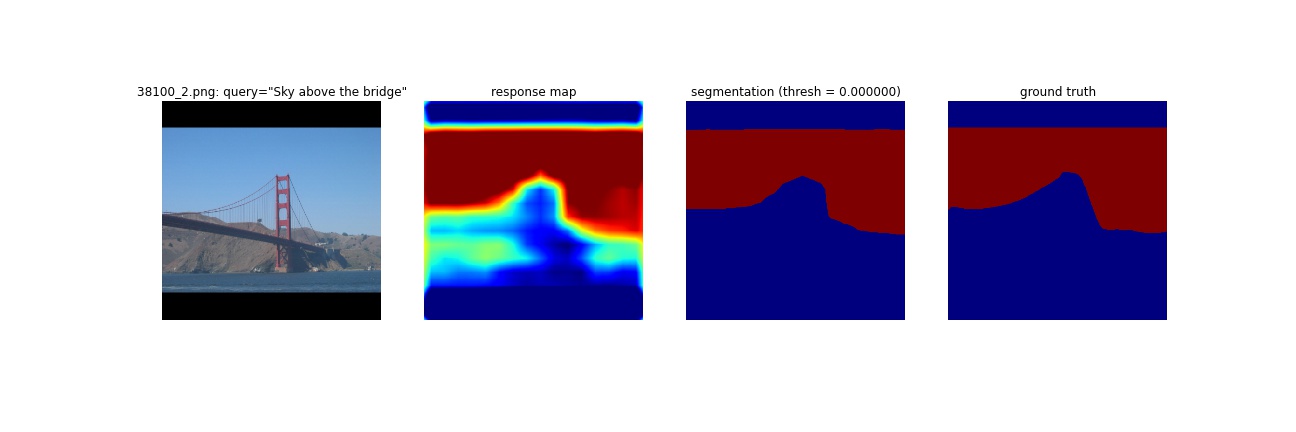} \\
\small{query expression=\refexp{water}} \\
\includegraphics[trim = 56mm 50mm 45mm 45mm, clip=true,width=0.90\textwidth,height=\textheight,keepaspectratio]{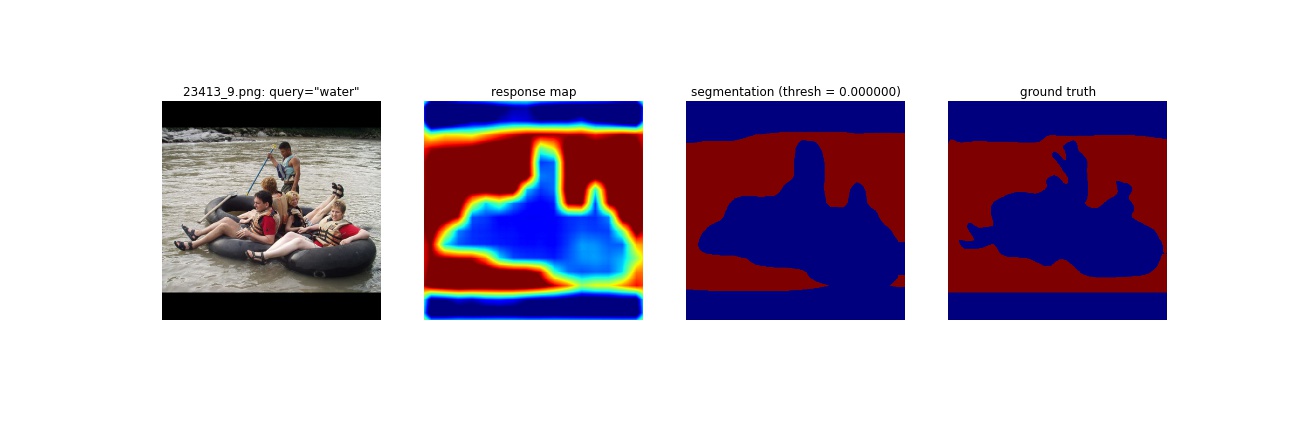} \\
\small{query expression=\refexp{wall above the people}} \\
\includegraphics[trim = 56mm 50mm 45mm 45mm, clip=true,width=0.90\textwidth,height=\textheight,keepaspectratio]{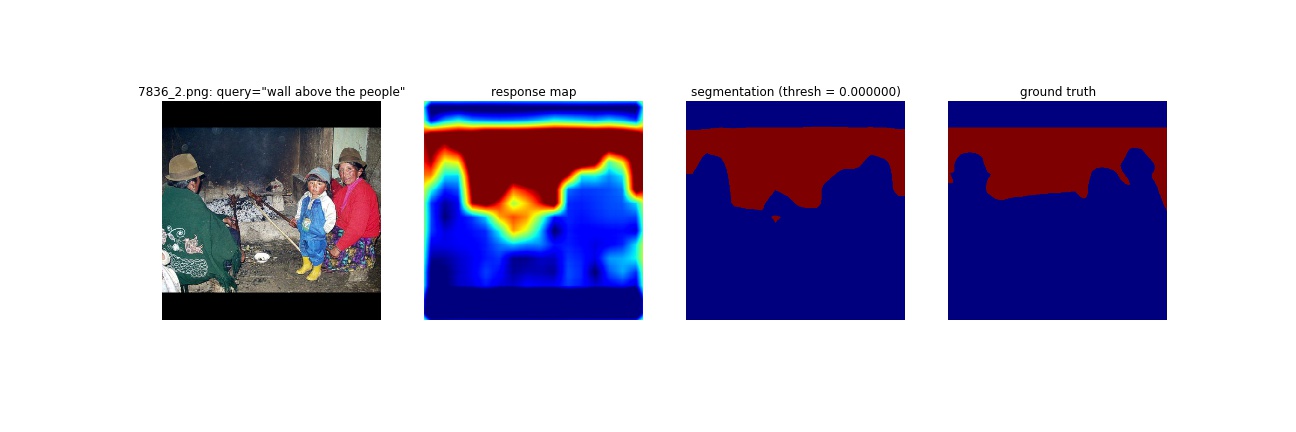} \\
\small{query expression=\refexp{the ground surrounding her}} \\
\includegraphics[trim = 56mm 38mm 45mm 35mm, clip=true,width=0.90\textwidth,height=\textheight,keepaspectratio]{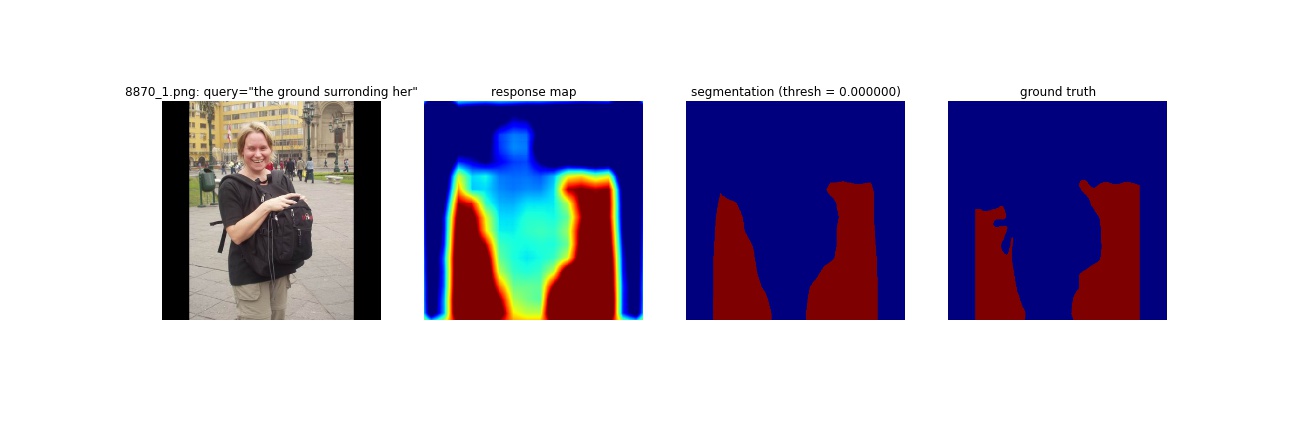} \\
\caption{Segmentation examples on stuff regions in the ReferIt dataset.}
\label{fig:sample_stuff}

\vspace{0.25 cm}

\begin{tabularx}{0.95\linewidth}{YYYY}
input image & response map & our prediction & ground-truth \\ \hline
\end{tabularx} \\
\small{query expression=\refexp{church}} \\
\includegraphics[trim = 56mm 50mm 45mm 45mm, clip=true,width=0.90\textwidth,height=\textheight,keepaspectratio]{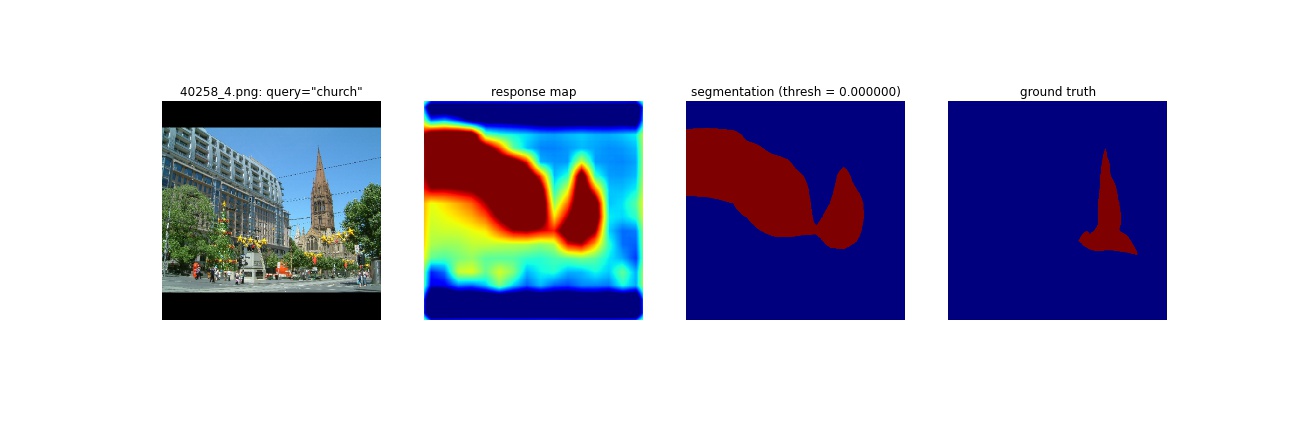} \\
\small{query expression=\refexp{right bird}} \\
\includegraphics[trim = 56mm 53mm 45mm 48mm, clip=true,width=0.90\textwidth,height=\textheight,keepaspectratio]{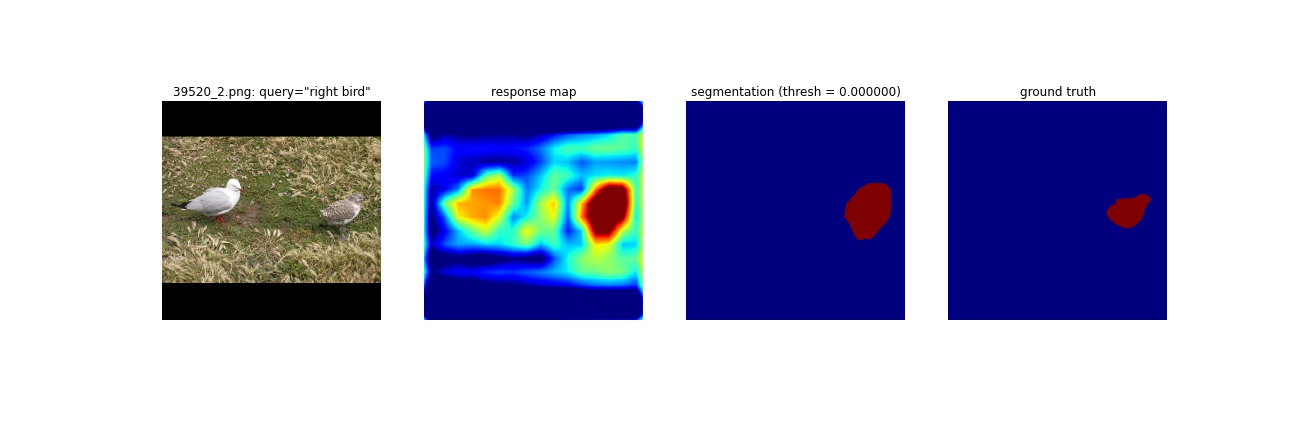} \\
\small{query expression=\refexp{plants below sign}} \\
\includegraphics[trim = 56mm 38mm 45mm 35mm, clip=true,width=0.90\textwidth,height=\textheight,keepaspectratio]{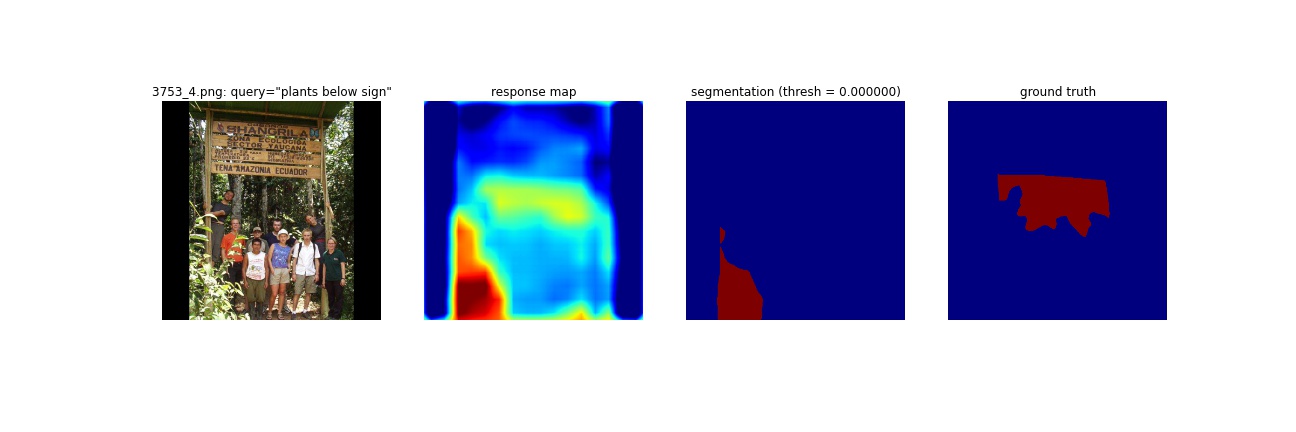} \\
\caption{Some failure cases where $\text{IoU} < 50\%$ between prediction and ground-truth.}
\label{fig:sample_failure}
\end{figure}

\begin{figure}[t]
\centering
\begin{tabularx}{0.95\linewidth}{YYYY}
input image & response map & our prediction & ground-truth \\ \hline
\end{tabularx} \\
\small{query expression=\refexp{greenery in foreground}} \\
\includegraphics[trim = 56mm 38mm 45mm 35mm, clip=true,width=0.90\textwidth,height=\textheight,keepaspectratio]{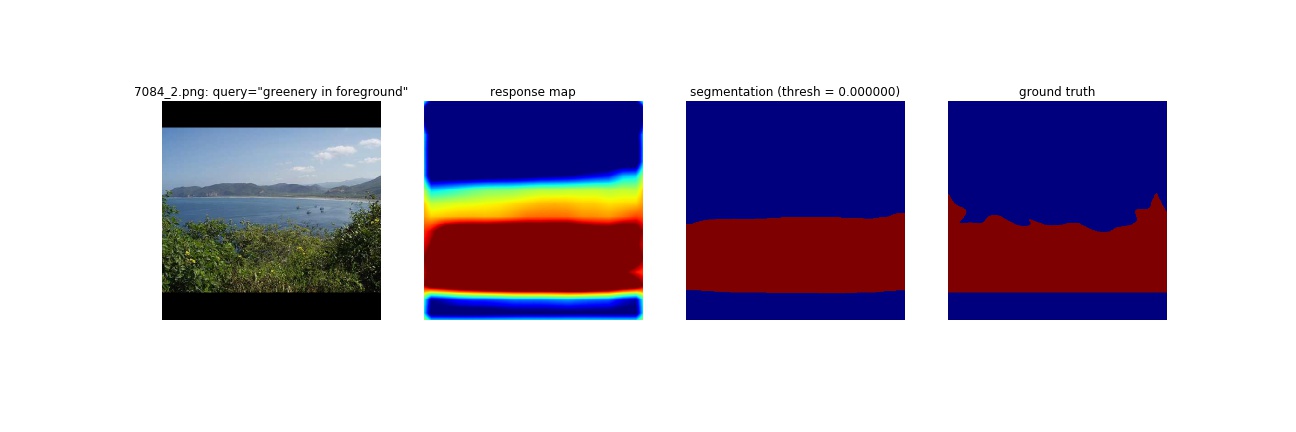} \\
\small{query expression=\refexp{sky}} \\
\includegraphics[trim = 56mm 38mm 45mm 35mm, clip=true,width=0.90\textwidth,height=\textheight,keepaspectratio]{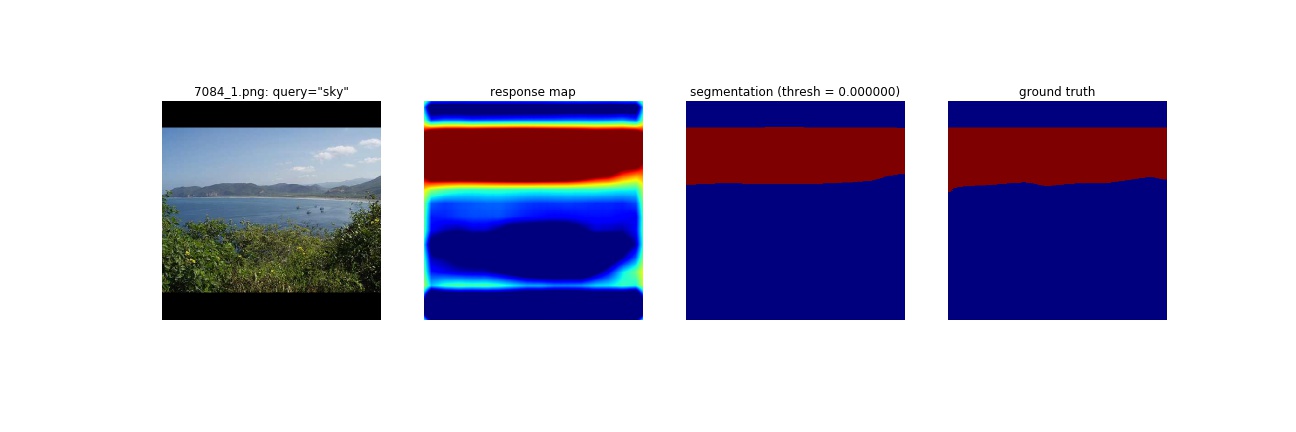} \\
\small{query expression=\refexp{the black sky in the middle}} \\
\includegraphics[trim = 56mm 38mm 45mm 35mm, clip=true,width=0.90\textwidth,height=\textheight,keepaspectratio]{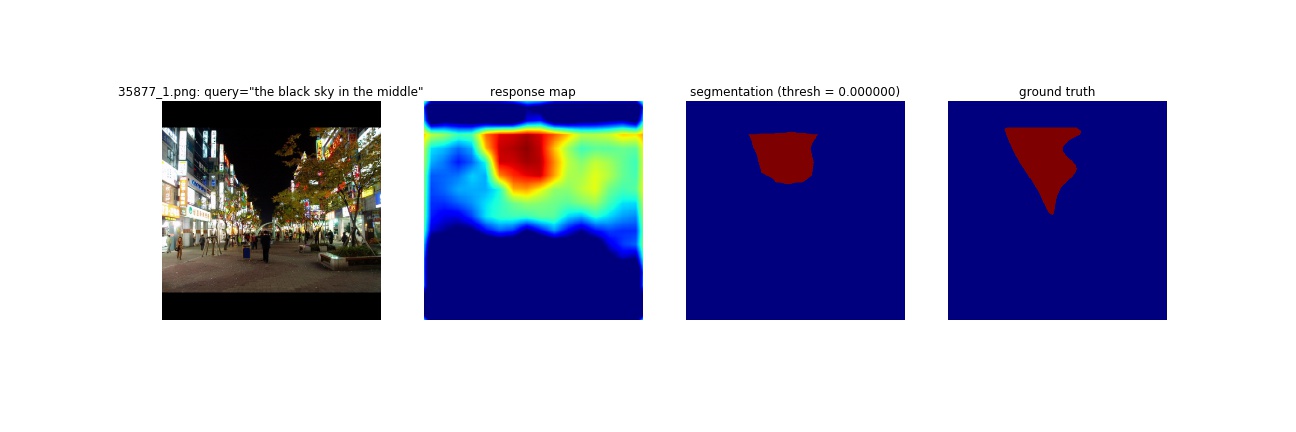} \\
\small{query expression=\refexp{bottom ground}} \\
\includegraphics[trim = 56mm 38mm 45mm 35mm, clip=true,width=0.90\textwidth,height=\textheight,keepaspectratio]{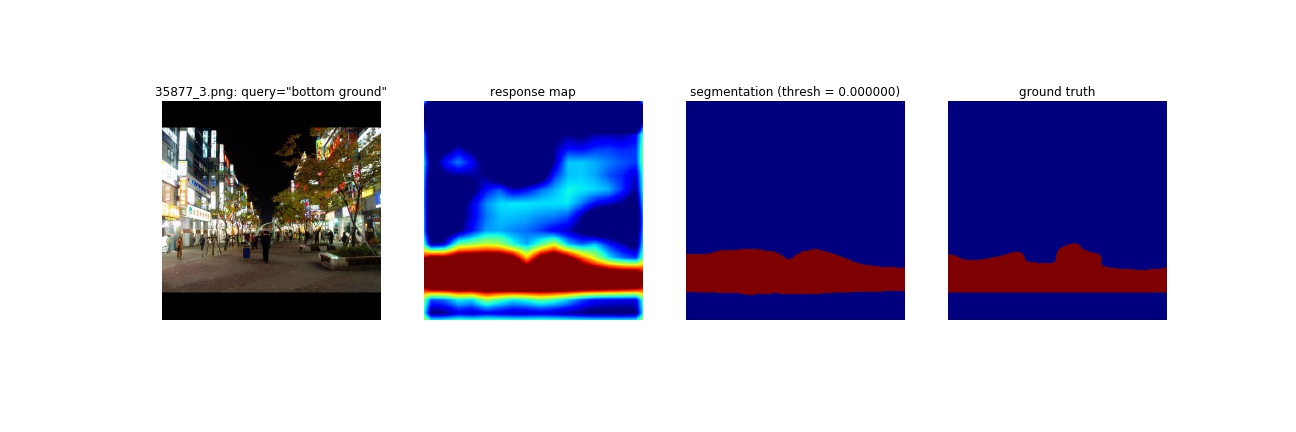} \\
\small{query expression=\refexp{city building}} \\
\includegraphics[trim = 56mm 38mm 45mm 35mm, clip=true,width=0.90\textwidth,height=\textheight,keepaspectratio]{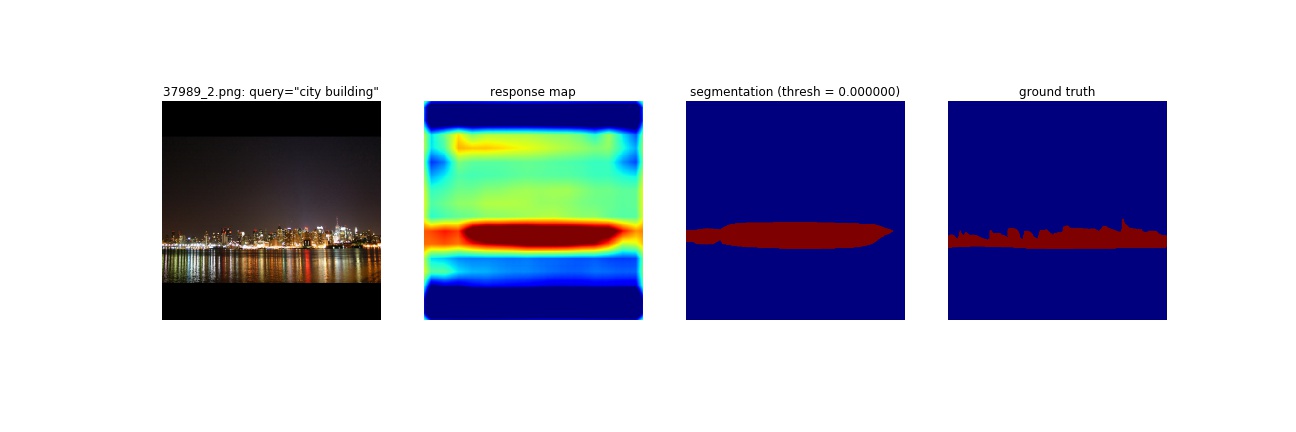} \\
\small{query expression=\refexp{water}} \\
\includegraphics[trim = 56mm 38mm 45mm 35mm, clip=true,width=0.90\textwidth,height=\textheight,keepaspectratio]{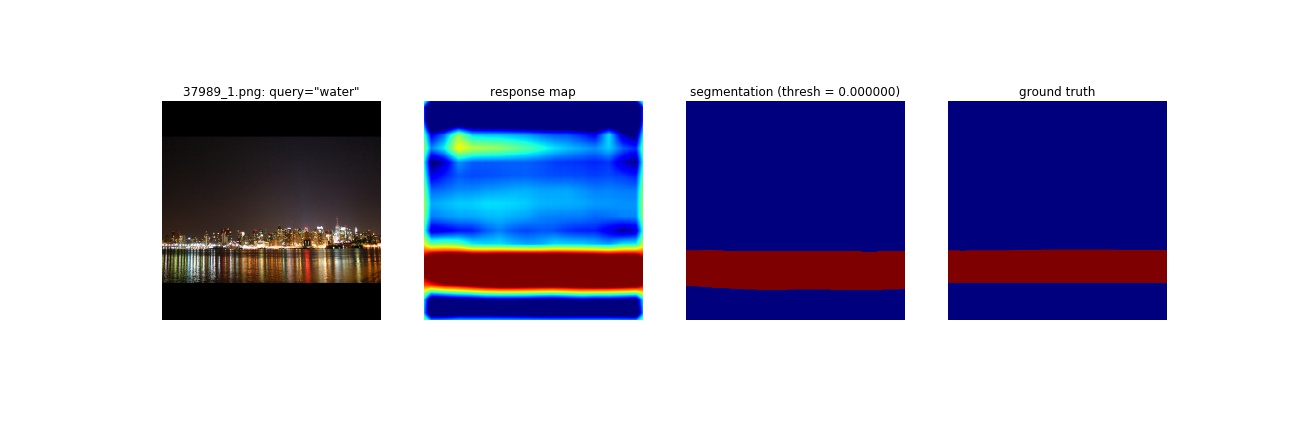} \\
\caption{Segmentation examples of different referential expressions on the same image from the ReferIt dataset.}
\label{fig:diff_exp_same_image}
\end{figure}

\begin{figure}[t]
\centering
\begin{tabularx}{0.95\linewidth}{YYYY}
input image & response map & our prediction & ground-truth \\ \hline
\end{tabularx} \\
\small{query expression=\refexp{plane}} \\
\includegraphics[trim = 56mm 38mm 45mm 35mm, clip=true,width=0.90\textwidth,height=\textheight,keepaspectratio]{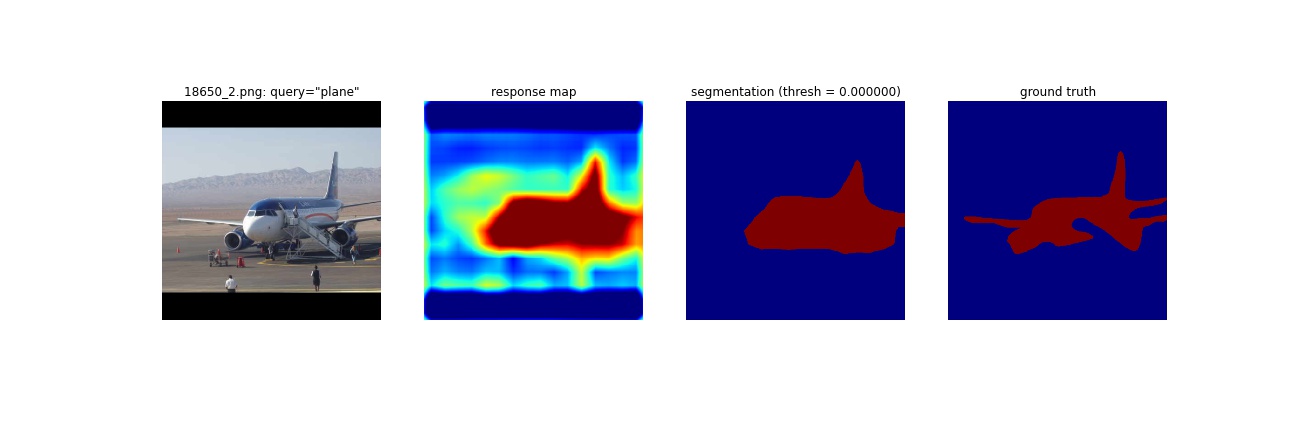} \\
\small{query expression=\refexp{animal in the tree}} \\
\includegraphics[trim = 56mm 38mm 45mm 35mm, clip=true,width=0.90\textwidth,height=\textheight,keepaspectratio]{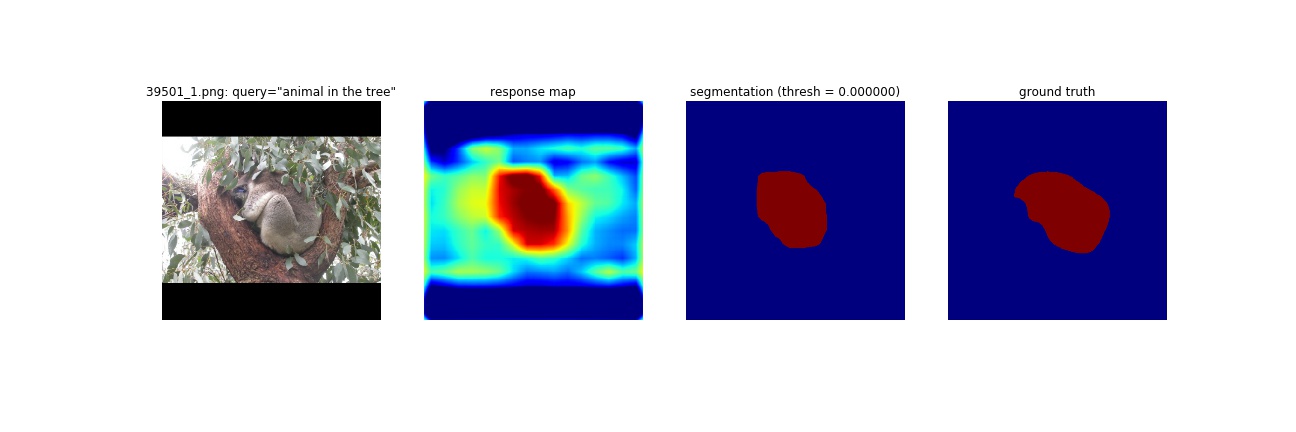} \\
\small{query expression=\refexp{woman on left}} \\
\includegraphics[trim = 56mm 38mm 45mm 35mm, clip=true,width=0.90\textwidth,height=\textheight,keepaspectratio]{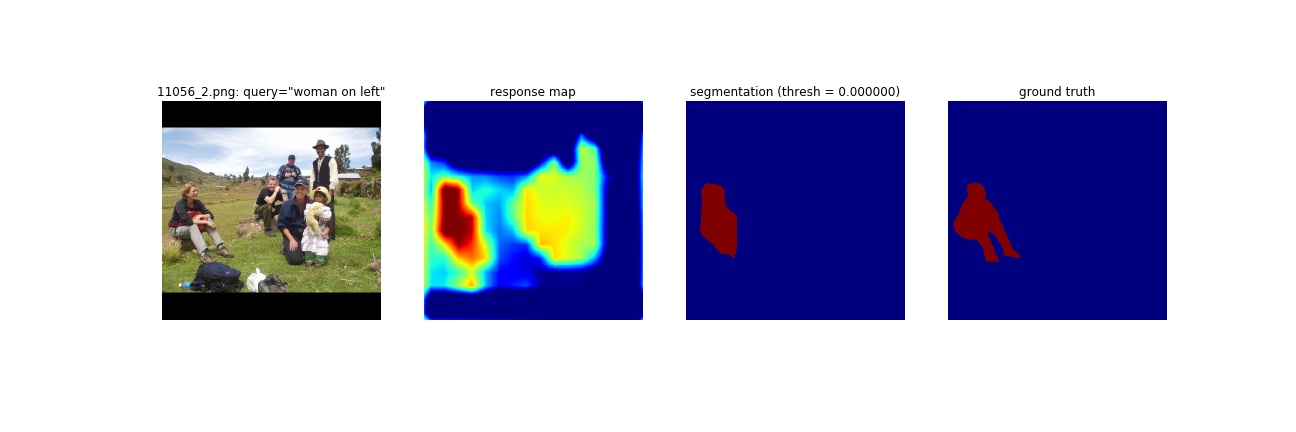} \\
\small{query expression=\refexp{squirrel}} \\
\includegraphics[trim = 56mm 38mm 45mm 35mm, clip=true,width=0.90\textwidth,height=\textheight,keepaspectratio]{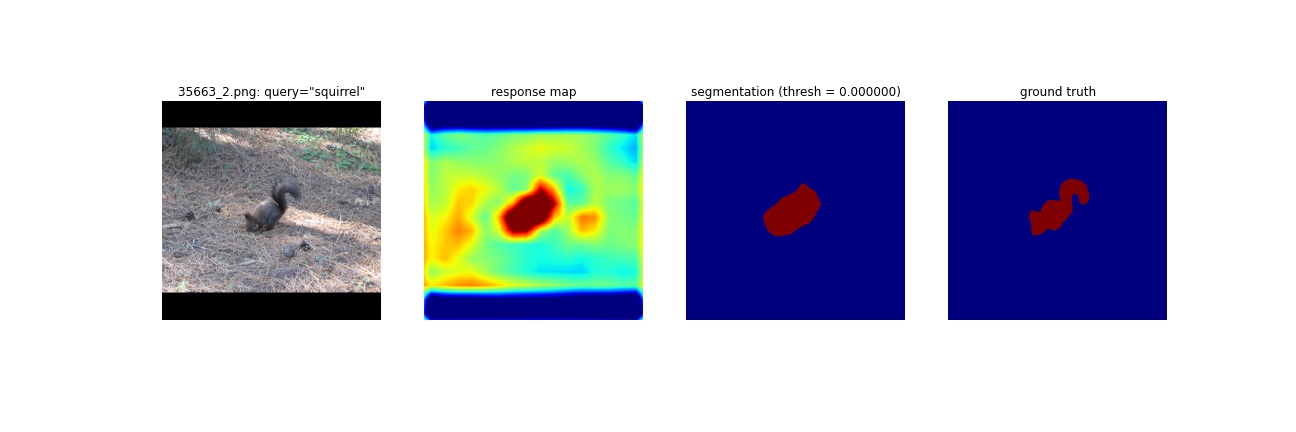} \\
\small{query expression=\refexp{left bed}} \\
\includegraphics[trim = 56mm 38mm 45mm 35mm, clip=true,width=0.90\textwidth,height=\textheight,keepaspectratio]{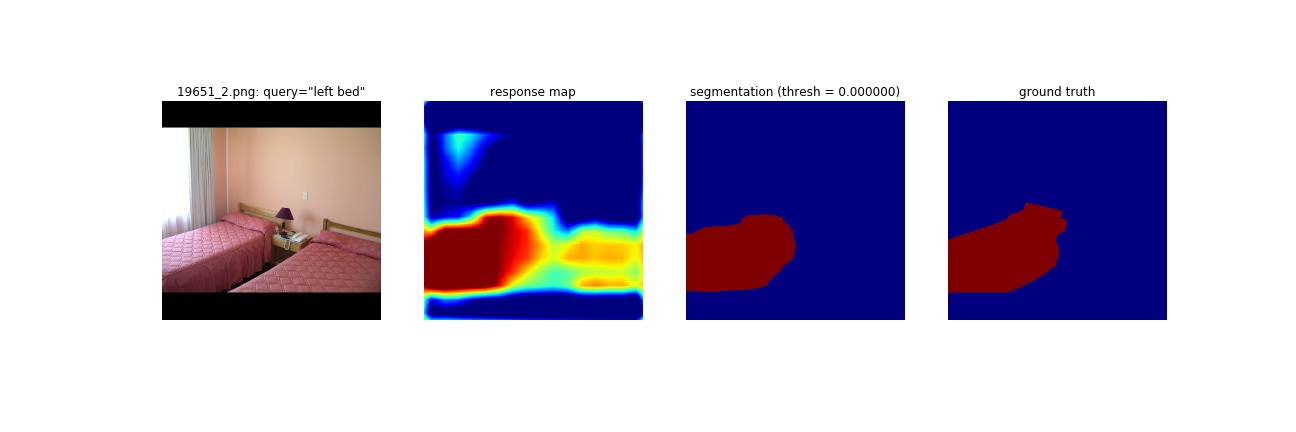} \\
\small{query expression=\refexp{church tower}} \\
\includegraphics[trim = 56mm 38mm 45mm 35mm, clip=true,width=0.90\textwidth,height=\textheight,keepaspectratio]{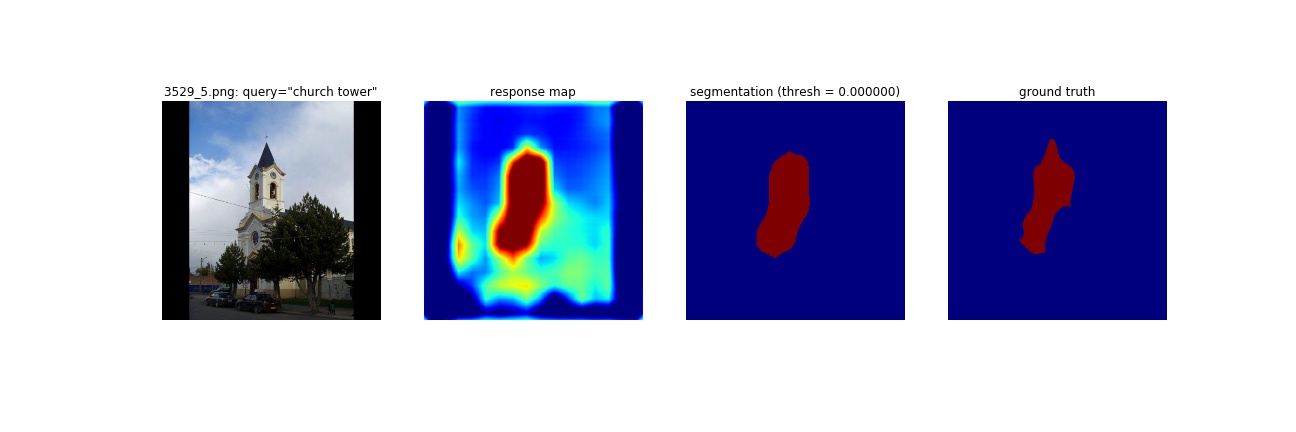} \\
\caption{More segmentation examples on object regions in the ReferIt dataset.}
\label{fig:sample_object_supp}
\end{figure}

\begin{figure}[t]
\centering
\begin{tabularx}{0.95\linewidth}{YYYY}
input image & response map & our prediction & ground-truth \\ \hline
\end{tabularx} \\
\small{query expression=\refexp{the water in the pool}} \\
\includegraphics[trim = 56mm 38mm 45mm 35mm, clip=true,width=0.90\textwidth,height=\textheight,keepaspectratio]{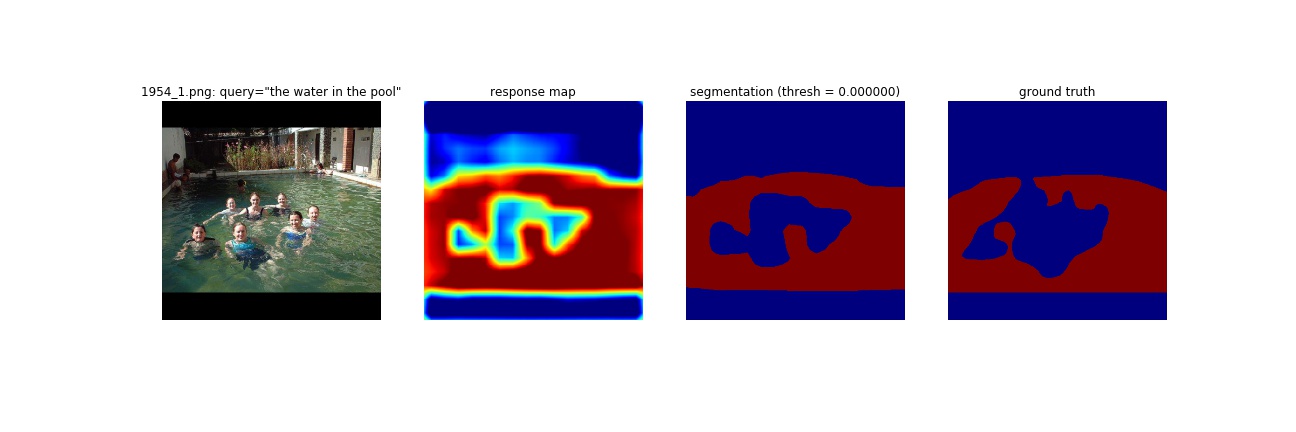} \\
\small{query expression=\refexp{ledge on left}} \\
\includegraphics[trim = 56mm 38mm 45mm 35mm, clip=true,width=0.90\textwidth,height=\textheight,keepaspectratio]{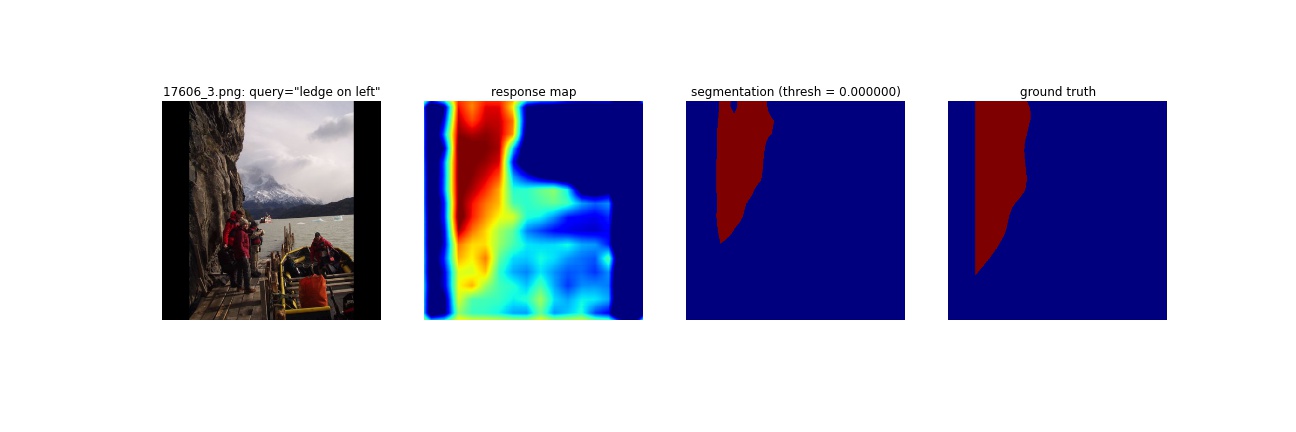} \\
\small{query expression=\refexp{dirt road on the left}} \\
\includegraphics[trim = 56mm 38mm 45mm 35mm, clip=true,width=0.90\textwidth,height=\textheight,keepaspectratio]{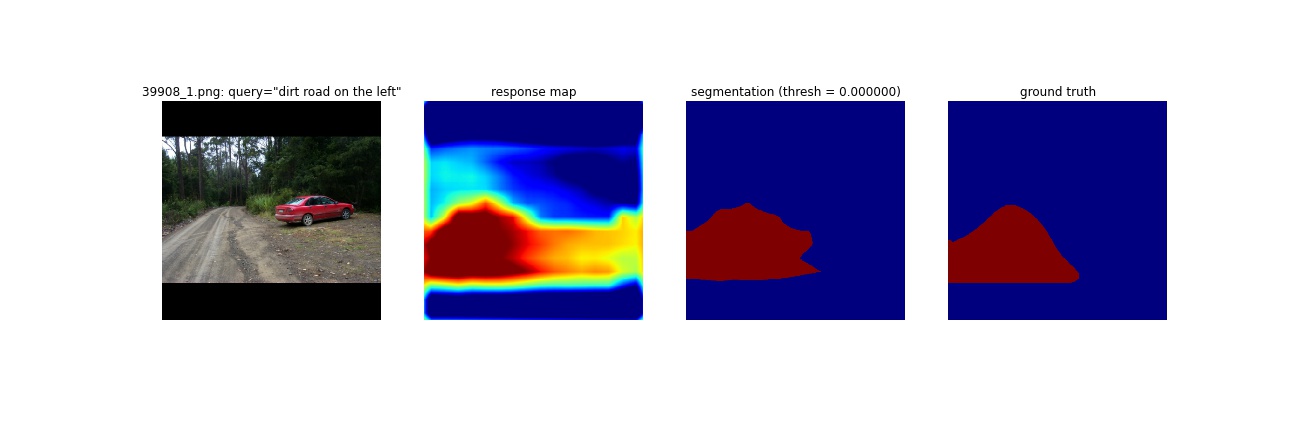} \\
\small{query expression=\refexp{anywhere in the diamond designs}} \\
\includegraphics[trim = 56mm 38mm 45mm 35mm, clip=true,width=0.90\textwidth,height=\textheight,keepaspectratio]{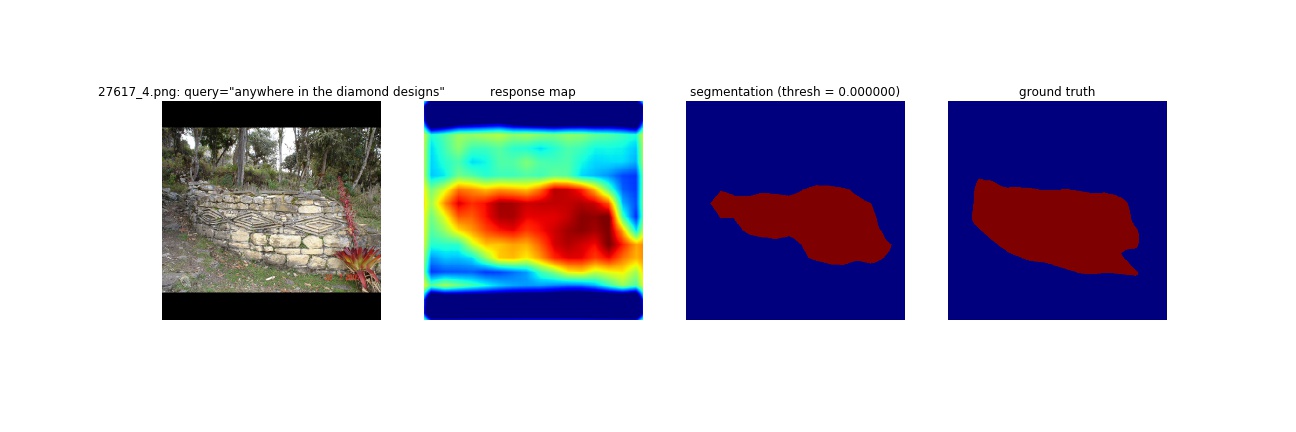} \\
\small{query expression=\refexp{grass right of people}} \\
\includegraphics[trim = 56mm 38mm 45mm 35mm, clip=true,width=0.90\textwidth,height=\textheight,keepaspectratio]{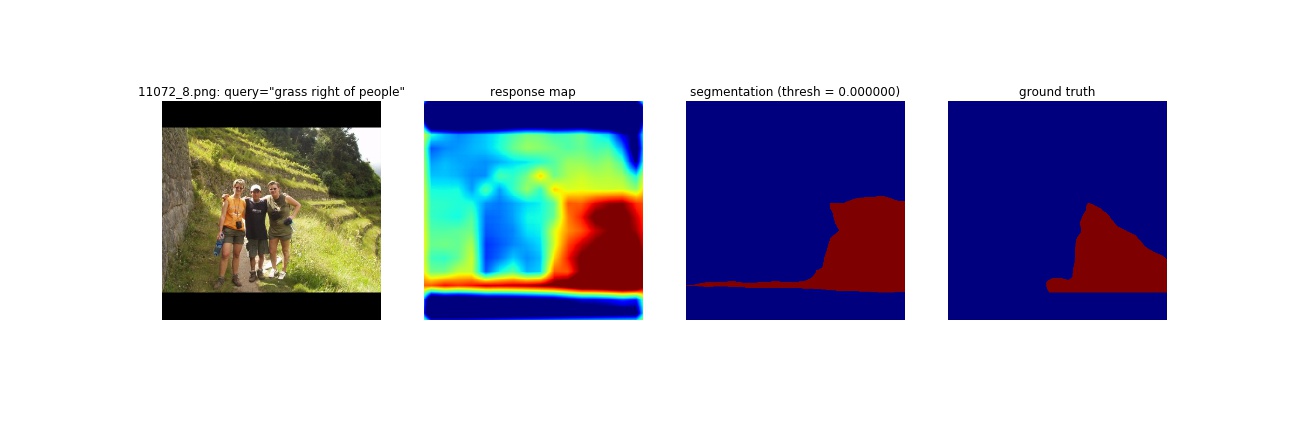} \\
\small{query expression=\refexp{sand between people on bottom}} \\
\includegraphics[trim = 56mm 38mm 45mm 35mm, clip=true,width=0.90\textwidth,height=\textheight,keepaspectratio]{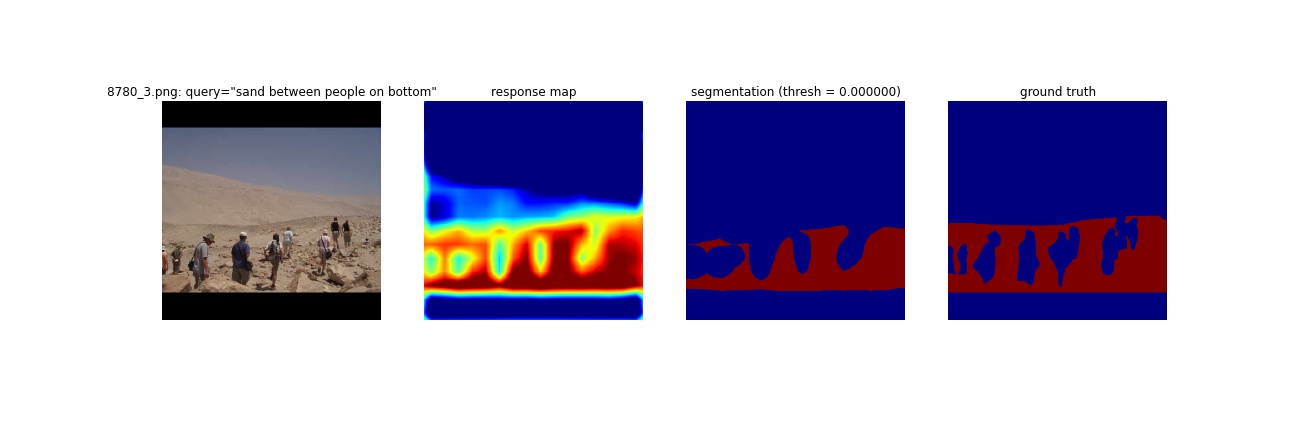} \\
\caption{More segmentation examples on stuff regions in the ReferIt dataset.}
\label{fig:sample_stuff_supp}
\end{figure}

The ReferIt dataset contains both object regions and stuff regions. Objects are those entities that have well-defined structures and closed boundaries, such as person, dog and airplane, while stuffs are those entities that do not have a fixed structure, such as sky, river, road and snow. Despite this difference, both object regions and stuff regions can be segmented through our model using the same approach. Figure \ref{fig:sample_object} shows some segmentation examples on object regions from our model, and Figure \ref{fig:sample_stuff} shows examples on stuff regions. It can be seen that our model can predict reasonable segmentation for both object expressions like ``bird on the left'' and stuff expressions like ``sky above the bridge''. Figure \ref{fig:diff_exp_same_image} visualizes some examples of different referential expressions on the same image. Figure \ref{fig:sample_object_supp} and Figure \ref{fig:sample_stuff_supp} show more segmentation examples on object and stuff regions.

\begin{figure}[t]
\centering
\begin{tabularx}{0.95\linewidth}{YYYY}
input image & response map & our prediction & ground-truth \\ \hline
\end{tabularx} \\
\small{query expression=\refexp{window above woman on the left}} \\
\includegraphics[trim = 56mm 38mm 45mm 35mm, clip=true,width=0.90\textwidth,height=\textheight,keepaspectratio]{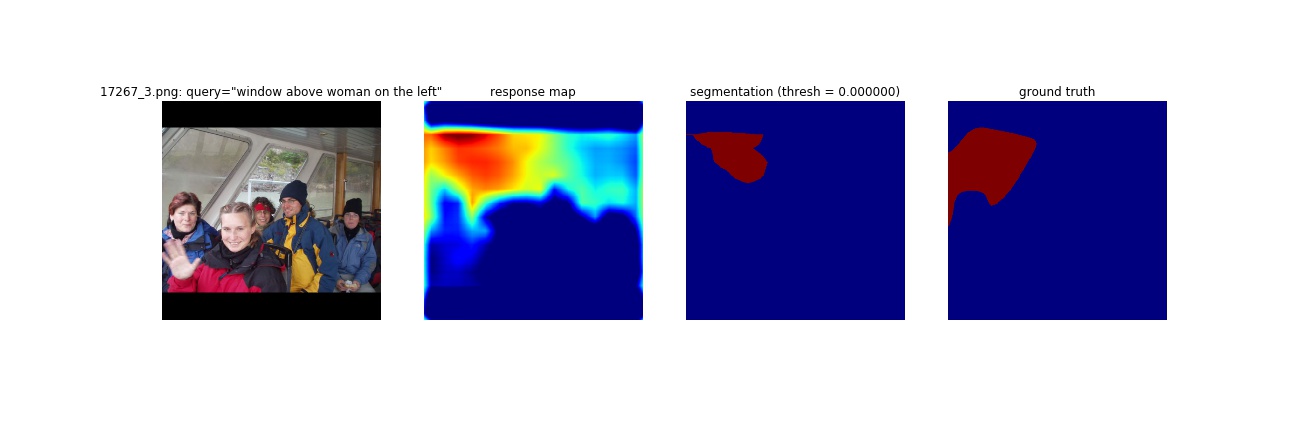} \\
\small{query expression=\refexp{man in the blue shorts by the railing}} \\
\includegraphics[trim = 56mm 38mm 45mm 35mm, clip=true,width=0.90\textwidth,height=\textheight,keepaspectratio]{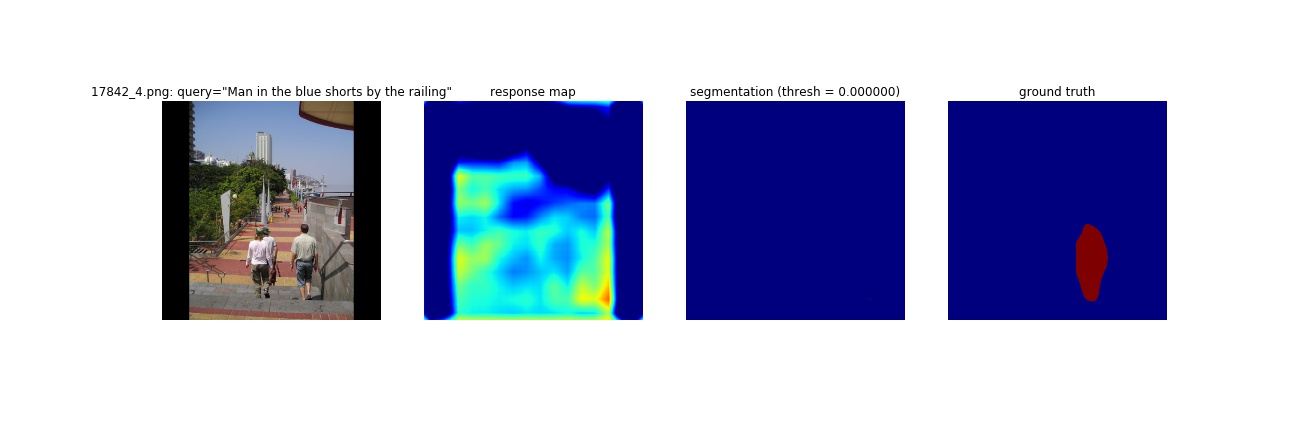} \\
\small{query expression=\refexp{yellow sign top right}} \\
\includegraphics[trim = 56mm 38mm 45mm 35mm, clip=true,width=0.90\textwidth,height=\textheight,keepaspectratio]{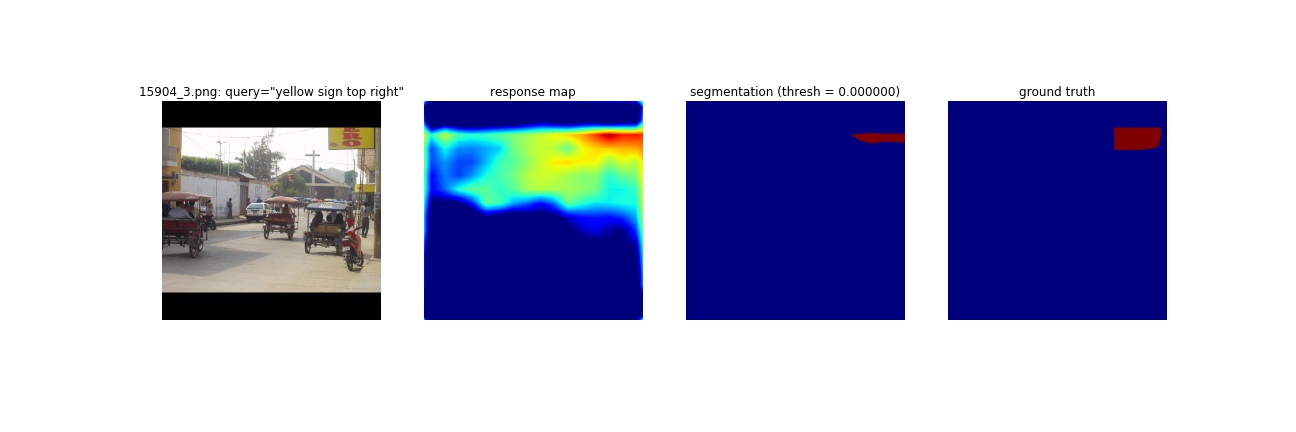} \\
\small{query expression=\refexp{rocks}} \\
\includegraphics[trim = 56mm 38mm 45mm 35mm, clip=true,width=0.90\textwidth,height=\textheight,keepaspectratio]{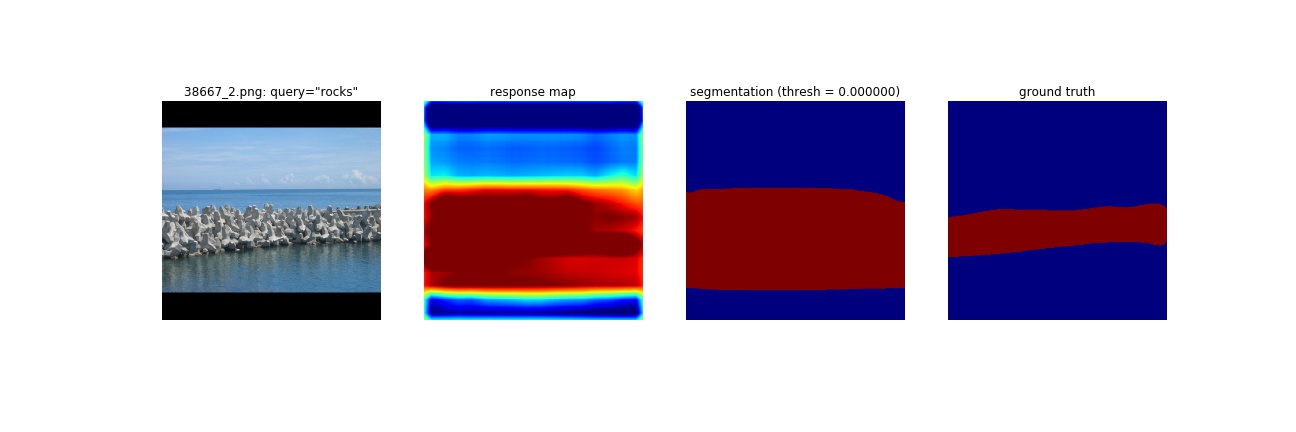} \\
\small{query expression=\refexp{leaf above the fruit}} \\
\includegraphics[trim = 56mm 38mm 45mm 35mm, clip=true,width=0.90\textwidth,height=\textheight,keepaspectratio]{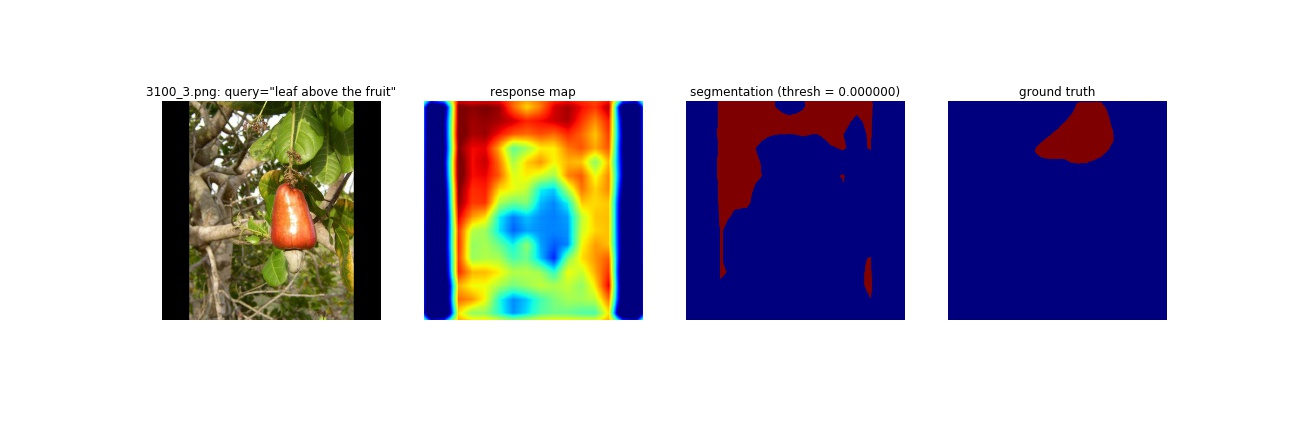} \\
\small{query expression=\refexp{sun}} \\
\includegraphics[trim = 56mm 38mm 45mm 35mm, clip=true,width=0.90\textwidth,height=\textheight,keepaspectratio]{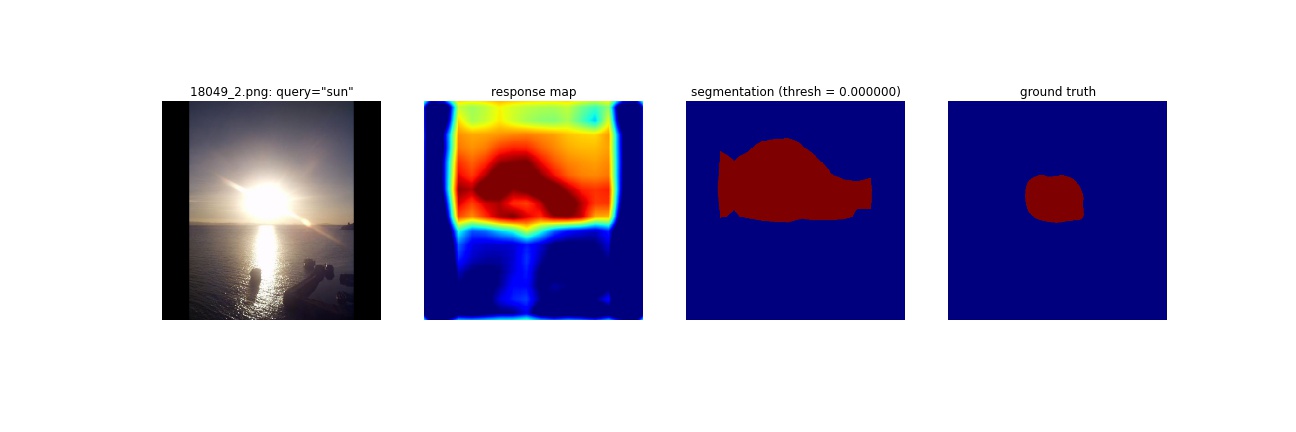} \\
\caption{More failure cases on the ReferIt dataset, where $\text{IoU} < 50\%$ between prediction and ground-truth.}
\label{fig:sample_failure_supp}
\end{figure}

Figure \ref{fig:sample_failure} shows some failure cases on the ReferIt dataset, where the IoU between prediction and ground-truth segmentation is less than 50\%. In some failure cases (\eg Figure \ref{fig:sample_failure}, middle), our model produces reasonable response maps that cover the target regions of the natural language referential expressions, but fails to precisely segment out the boundary of objects or stuffs. Figure \ref{fig:sample_failure_supp} shows more failure cases.

\begin{table}[t]
\begin{center}
\begin{tabularx}{0.9\textwidth}{YYYYYY}
\hline
\textbf{Method} & per-word & SCRC \cite{hu2015natural} grabcut & GroundeR \cite{rohrbach2015grounding} grabcut & MCG classification & Ours (high resolution) \\
\hline
time (sec) & 0.169 & 4.319 & 3.753 & 9.375 & 0.325 \\
\hline
\end{tabularx} \\
\end{center}
\caption{Average time consumption to segmentation an input (a given image and a natural language expression) using different methods.}
\label{tab:speed}
\end{table}

\textbf{Speed.}
We also compare the speed of our method and baseline methods. Table \ref{tab:speed} shows the average time consumption for different models to predict a segmentation at test time, on a single machine with NVIDIA Tesla K40 GPU. It can be seen that although our method is slower than the per-word segmentation baseline, it is significantly faster than proposal-based methods such as ``SCRC grabcut'' or ``MCG classification''.

\section{Conclusion}

In this paper, we address the challenging problem of segmenting natural language expressions, to generate a pixelwise segmentation output for the image region described by the referential expression. To solve this problem, we propose an end-to-end trainable recurrent convolutional neural network model to encode the expression into a vector representation, extract a spatial feature map representation from the image, and output pixelwise segmentation based on fully convolutional classifier and upsampling. Our model can efficiently predict segmentation output for referential expressions that describe single or multiple objects or stuffs. Experimental results on a benchmark dataset demonstrate that our model outperforms baseline methods by a large margin.

\section*{Acknowledgments}

The authors are grateful to Lisa Hendricks and Marcel Simon for feedback on drafts. This work was supported by DARPA, AFRL, DoD MURI award N000141110688, NSF awards IIS-1427425 and IIS-1212798, and the Berkeley Vision and Learning Center. Marcus Rohrbach was supported by a fellowship within the FITweltweit-Program of the German Academic Exchange Service (DAAD).

\bibliographystyle{splncs}
\bibliography{text_objseg_arxiv}

\begin{figure}[t]
\centering
\begin{tabularx}{0.95\linewidth}{YYYY}
input image & response map & our prediction & ground-truth \\ \hline
\end{tabularx} \\
\small{query expression=\refexp{group of people}} \\
\includegraphics[trim = 56mm 38mm 45mm 35mm, clip=true,width=0.90\textwidth,height=\textheight,keepaspectratio]{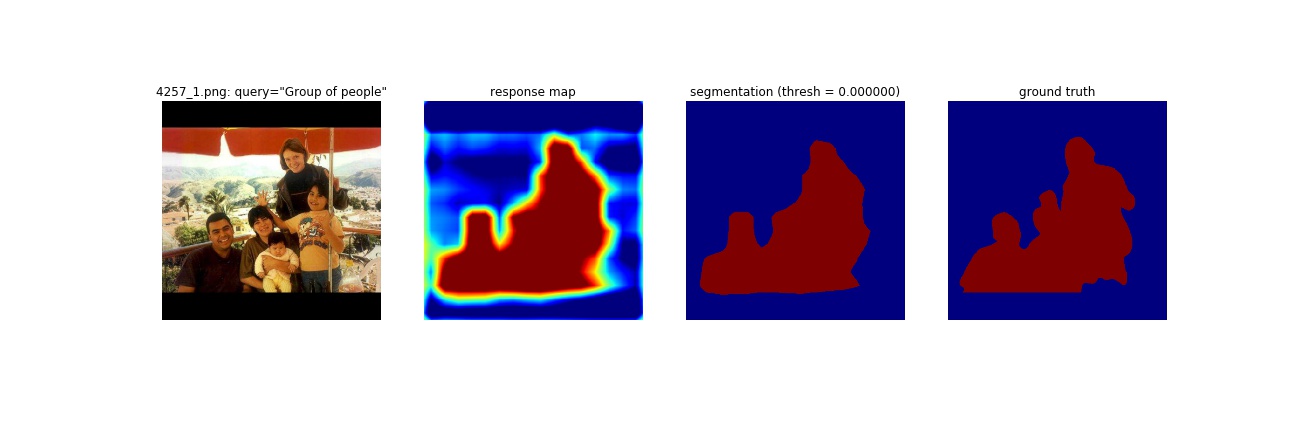} \\
\small{query expression=\refexp{umbrella}} \\
\includegraphics[trim = 56mm 38mm 45mm 35mm, clip=true,width=0.90\textwidth,height=\textheight,keepaspectratio]{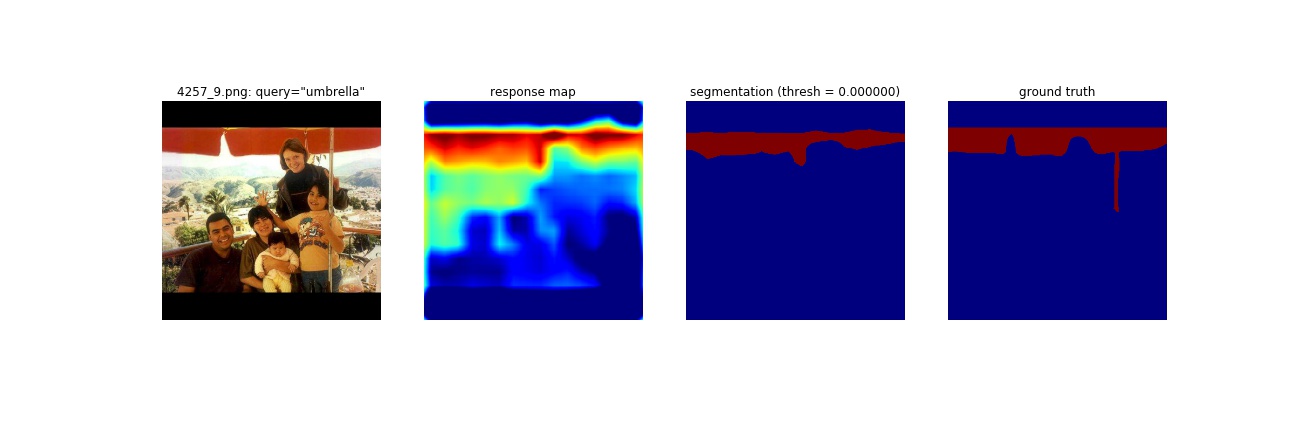} \\
\small{query expression=\refexp{three people right}} \\
\includegraphics[trim = 56mm 38mm 45mm 35mm, clip=true,width=0.90\textwidth,height=\textheight,keepaspectratio]{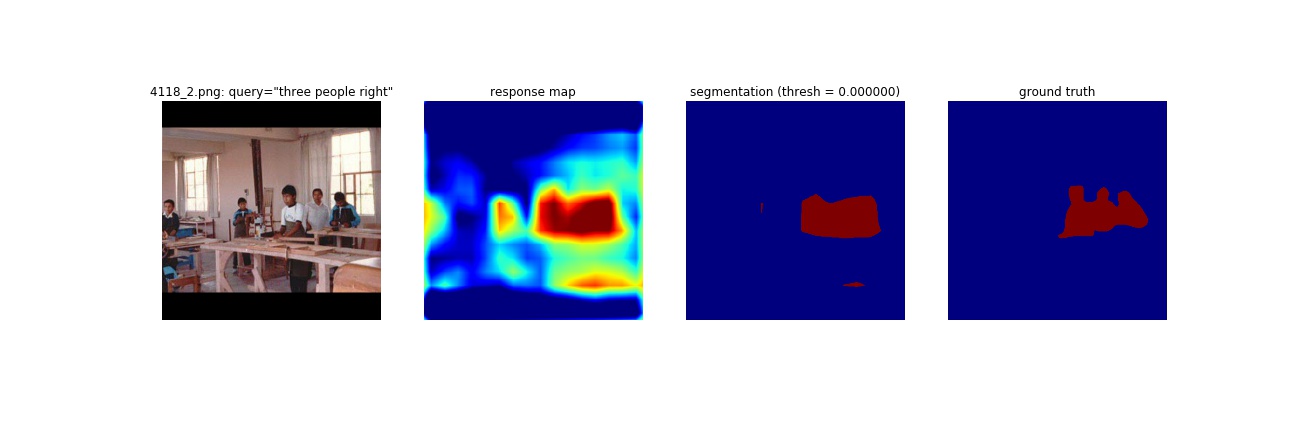} \\
\small{query expression=\refexp{person on left}} \\
\includegraphics[trim = 56mm 38mm 45mm 35mm, clip=true,width=0.90\textwidth,height=\textheight,keepaspectratio]{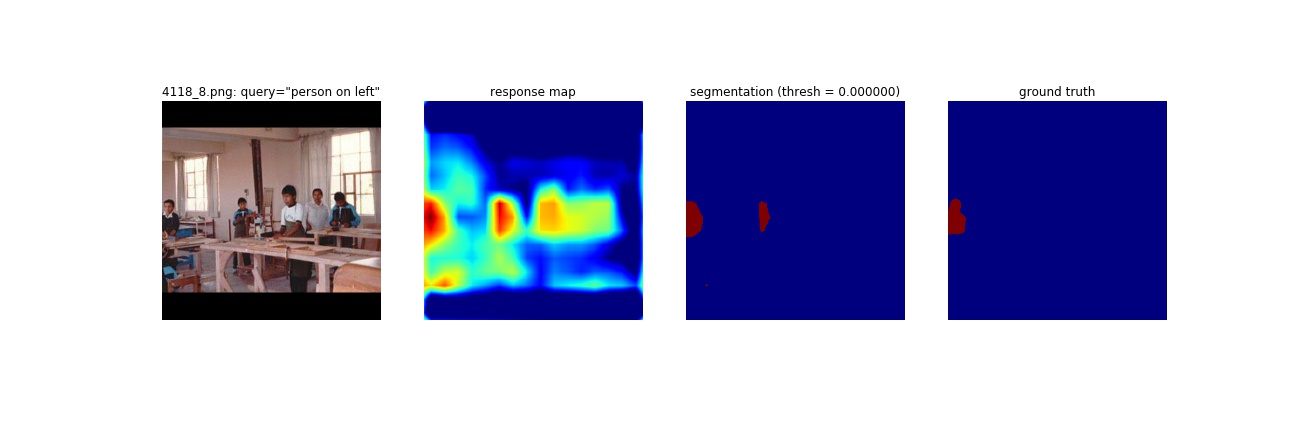} \\
\small{query expression=\refexp{trees on right}} \\
\includegraphics[trim = 56mm 38mm 45mm 35mm, clip=true,width=0.90\textwidth,height=\textheight,keepaspectratio]{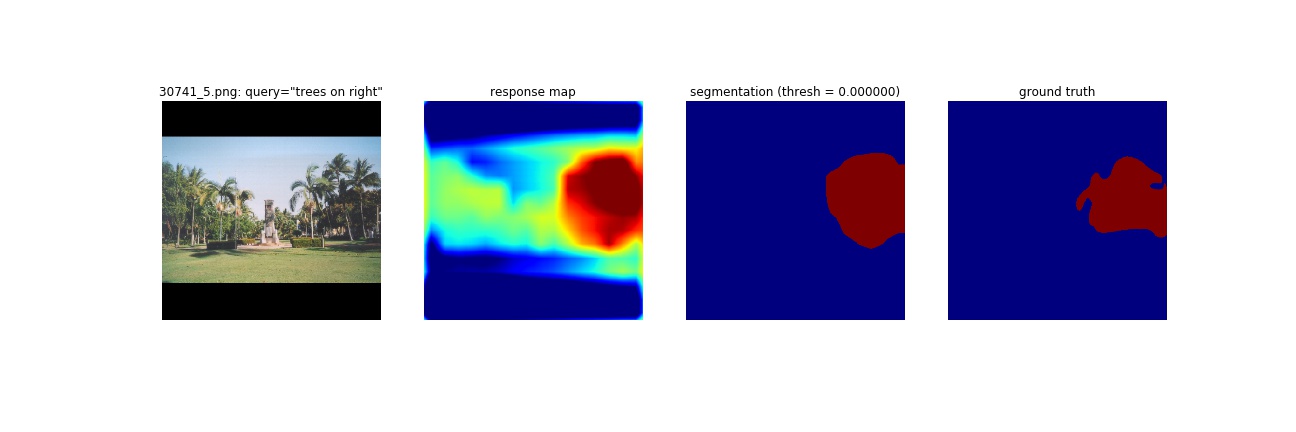} \\
\small{query expression=\refexp{trees on left side}} \\
\includegraphics[trim = 56mm 38mm 45mm 35mm, clip=true,width=0.90\textwidth,height=\textheight,keepaspectratio]{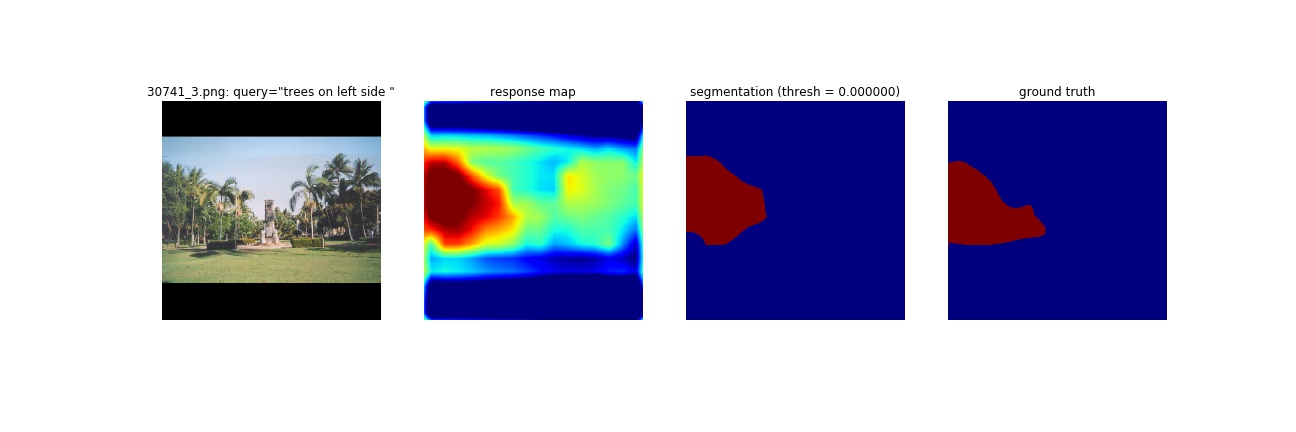} \\
\end{figure}

\begin{figure}[t]
\centering
\begin{tabularx}{0.95\linewidth}{YYYY}
input image & response map & our prediction & ground-truth \\ \hline
\end{tabularx} \\
\small{query expression=\refexp{dude on left}} \\
\includegraphics[trim = 56mm 38mm 45mm 35mm, clip=true,width=0.90\textwidth,height=\textheight,keepaspectratio]{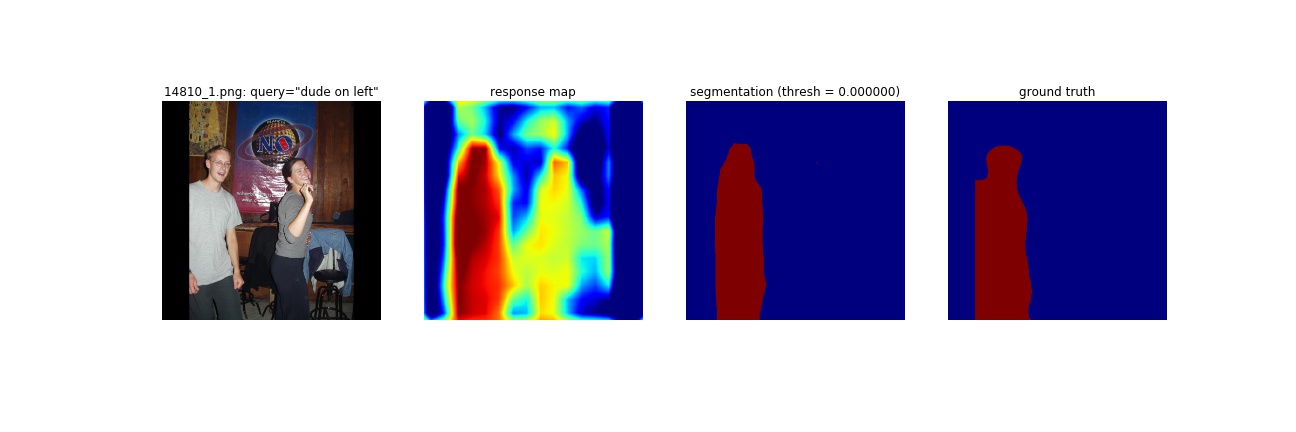} \\
\small{query expression=\refexp{any one on boat}} \\
\includegraphics[trim = 56mm 38mm 45mm 35mm, clip=true,width=0.90\textwidth,height=\textheight,keepaspectratio]{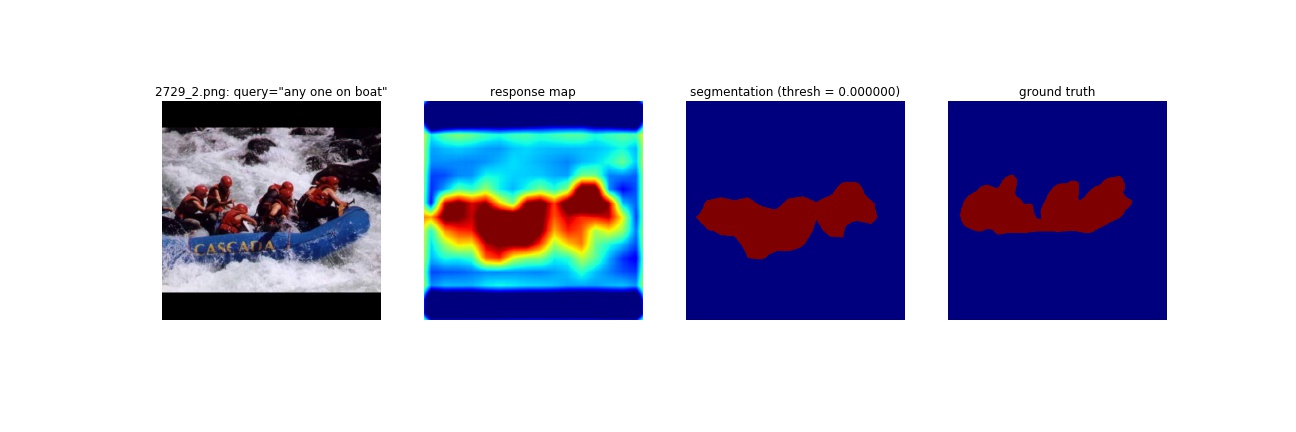} \\
\small{query expression=\refexp{white horse, left}} \\
\includegraphics[trim = 56mm 38mm 45mm 35mm, clip=true,width=0.90\textwidth,height=\textheight,keepaspectratio]{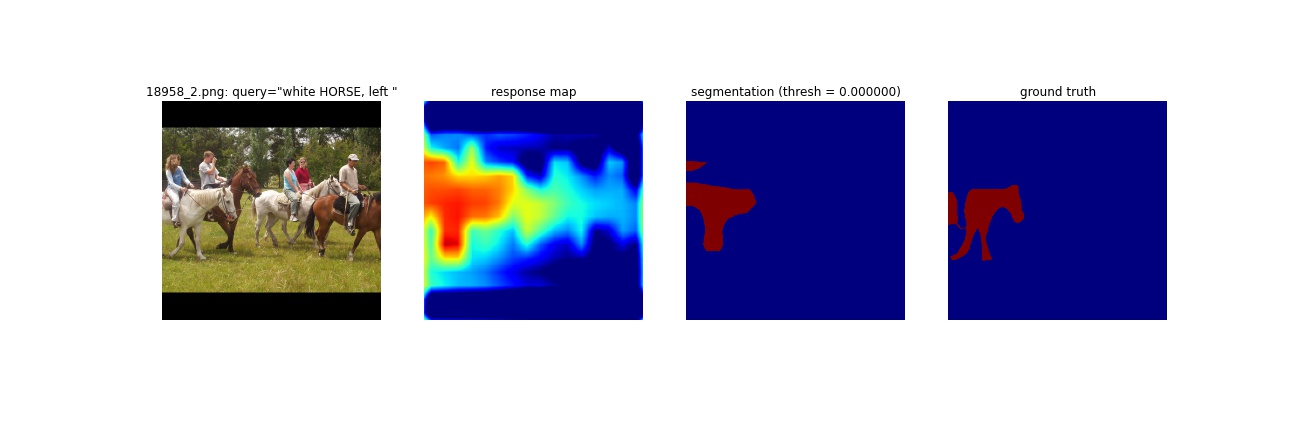} \\
\small{query expression=\refexp{the people in the middle}} \\
\includegraphics[trim = 56mm 38mm 45mm 35mm, clip=true,width=0.90\textwidth,height=\textheight,keepaspectratio]{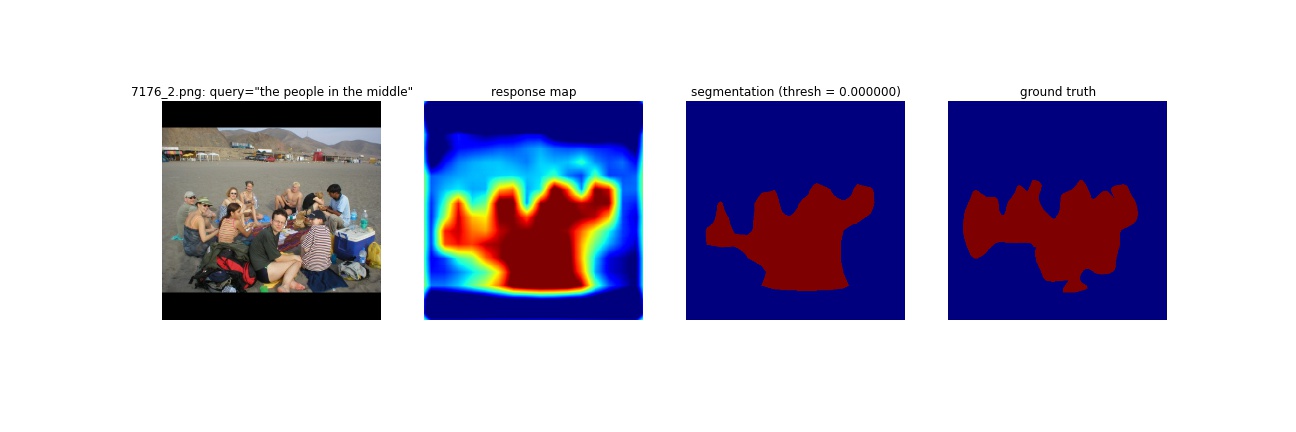} \\
\small{query expression=\refexp{the cactus on the right with 2 arms}} \\
\includegraphics[trim = 56mm 38mm 45mm 35mm, clip=true,width=0.90\textwidth,height=\textheight,keepaspectratio]{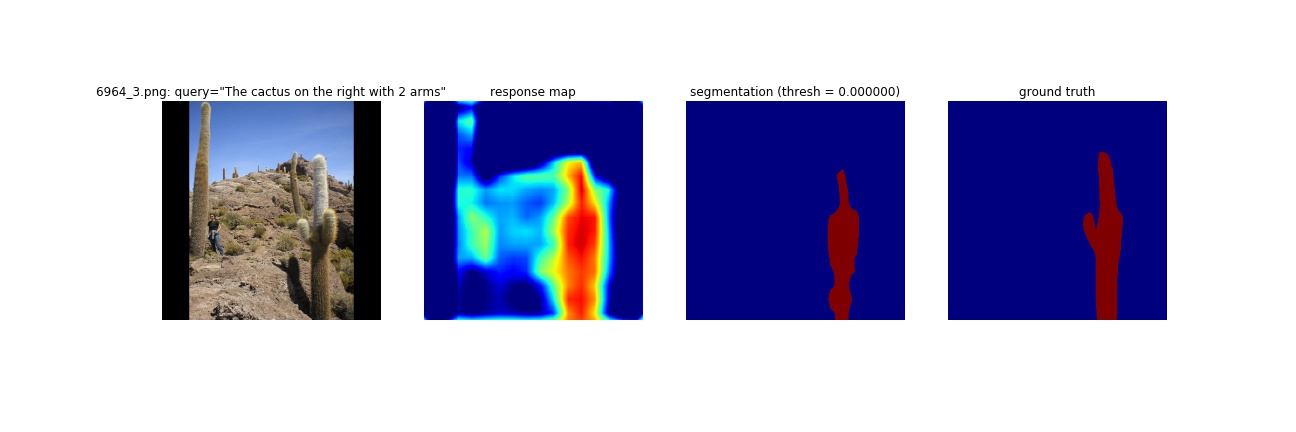} \\
\small{query expression=\refexp{llama left}} \\
\includegraphics[trim = 56mm 38mm 45mm 35mm, clip=true,width=0.90\textwidth,height=\textheight,keepaspectratio]{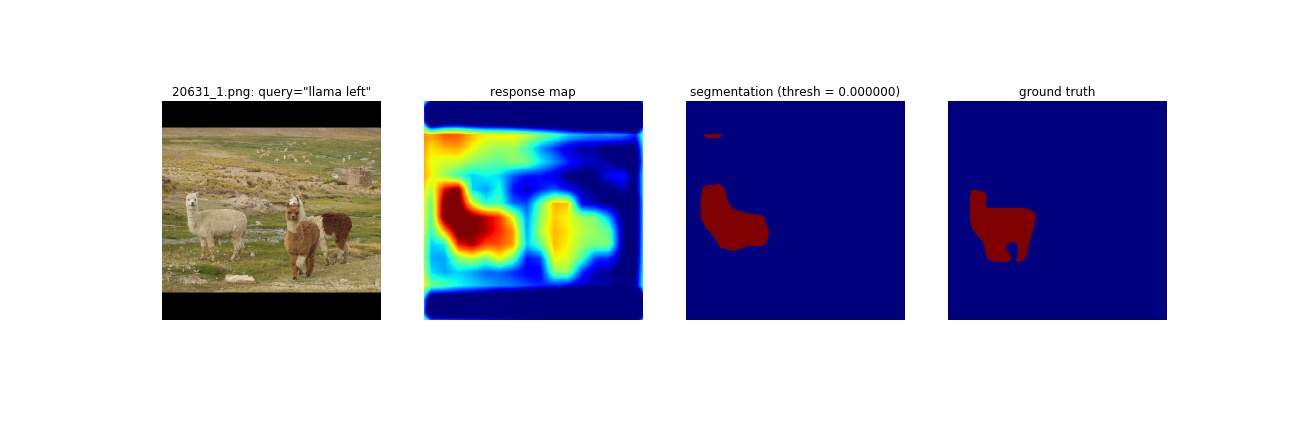} \\
\end{figure}

\begin{figure}[t]
\centering
\begin{tabularx}{0.95\linewidth}{YYYY}
input image & response map & our prediction & ground-truth \\ \hline
\end{tabularx} \\
\small{query expression=\refexp{bike wheel}} \\
\includegraphics[trim = 56mm 38mm 45mm 35mm, clip=true,width=0.90\textwidth,height=\textheight,keepaspectratio]{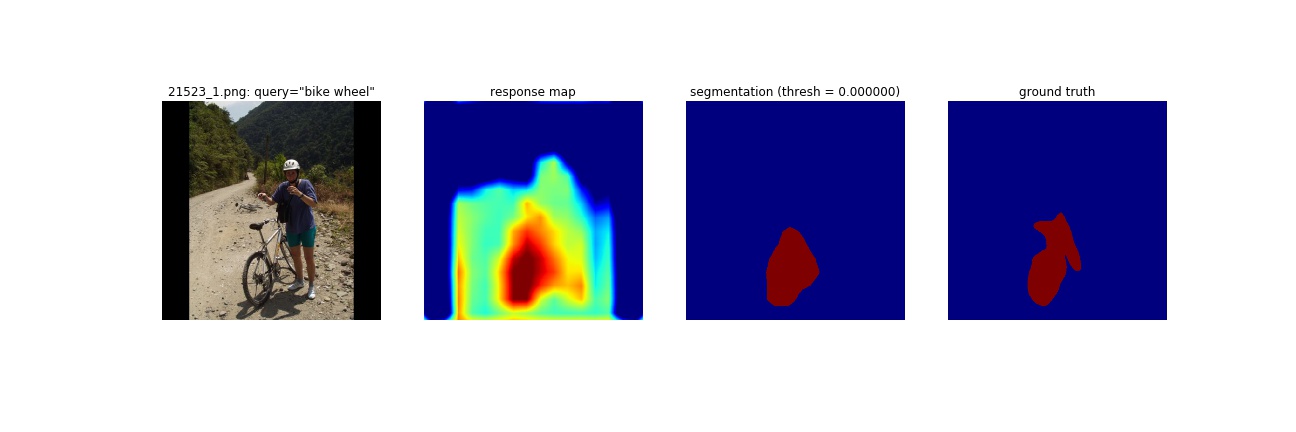} \\
\small{query expression=\refexp{hat}} \\
\includegraphics[trim = 56mm 38mm 45mm 35mm, clip=true,width=0.90\textwidth,height=\textheight,keepaspectratio]{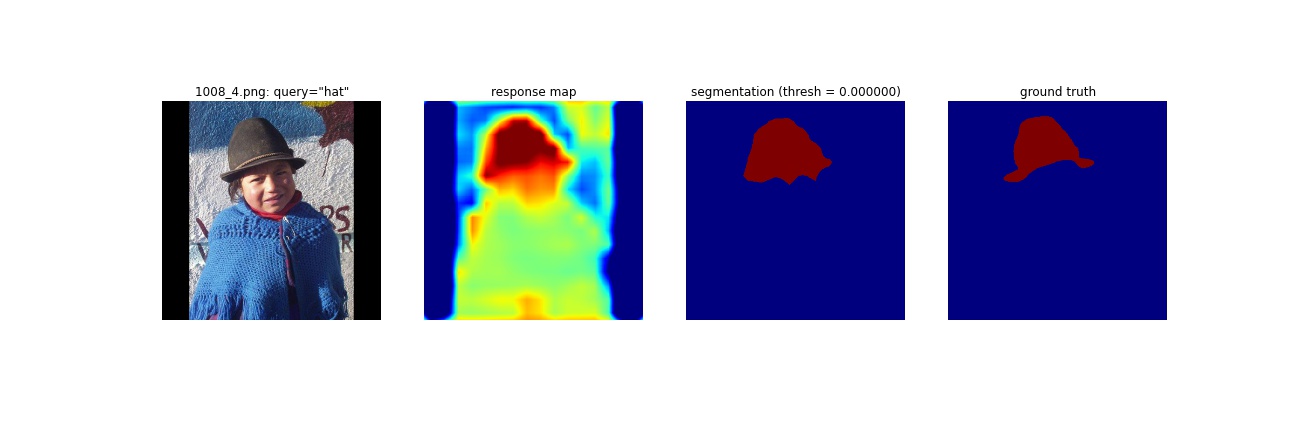} \\
\small{query expression=\refexp{people farthest on the right}} \\
\includegraphics[trim = 56mm 38mm 45mm 35mm, clip=true,width=0.90\textwidth,height=\textheight,keepaspectratio]{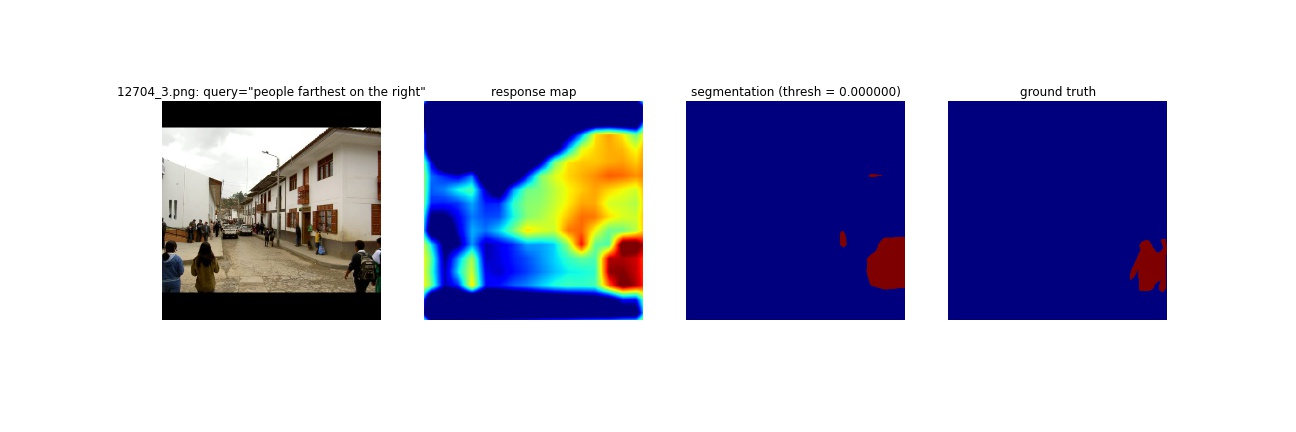} \\
\small{query expression=\refexp{angel}} \\
\includegraphics[trim = 56mm 38mm 45mm 35mm, clip=true,width=0.90\textwidth,height=\textheight,keepaspectratio]{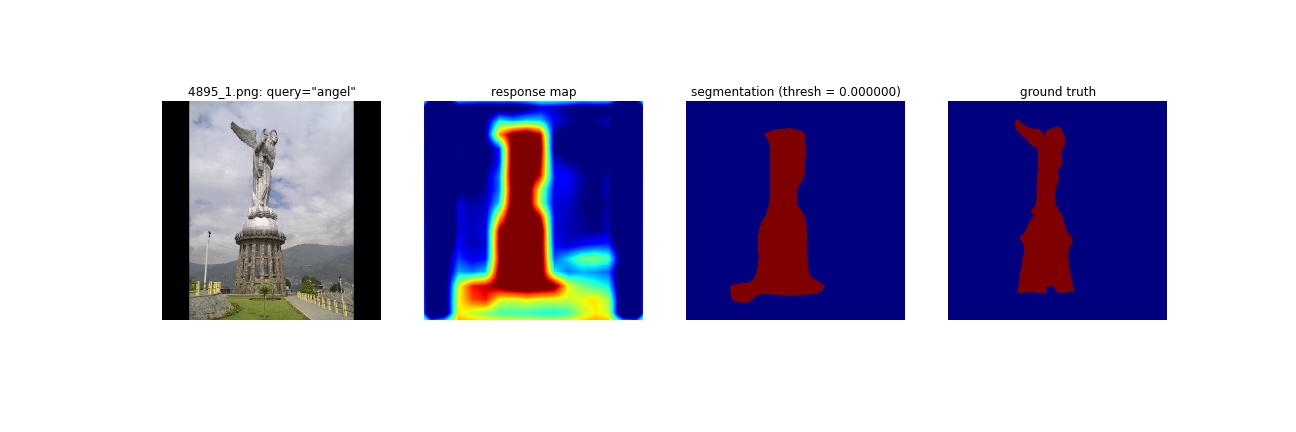} \\
\small{query expression=\refexp{biker}} \\
\includegraphics[trim = 56mm 38mm 45mm 35mm, clip=true,width=0.90\textwidth,height=\textheight,keepaspectratio]{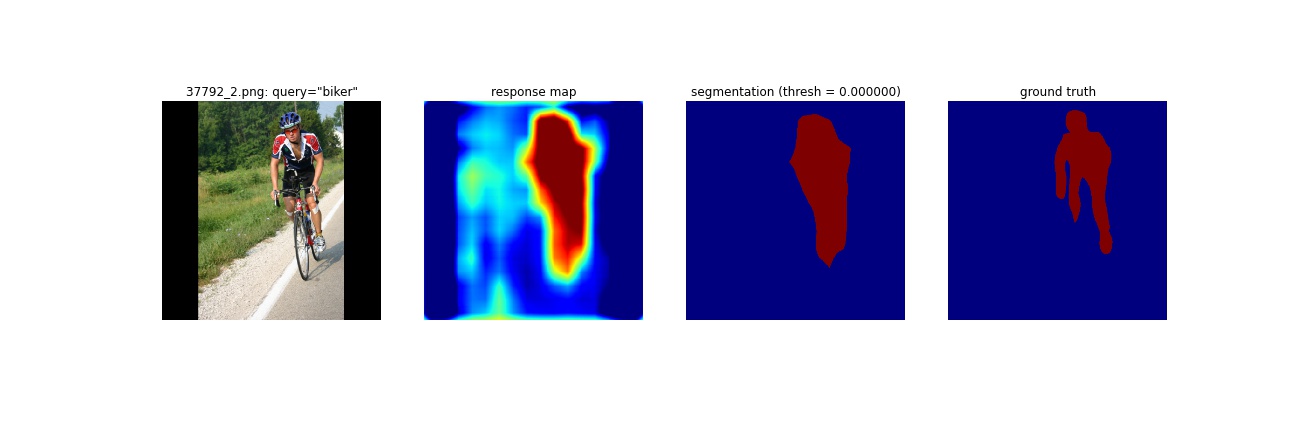} \\
\small{query expression=\refexp{two people on left}} \\
\includegraphics[trim = 56mm 38mm 45mm 35mm, clip=true,width=0.90\textwidth,height=\textheight,keepaspectratio]{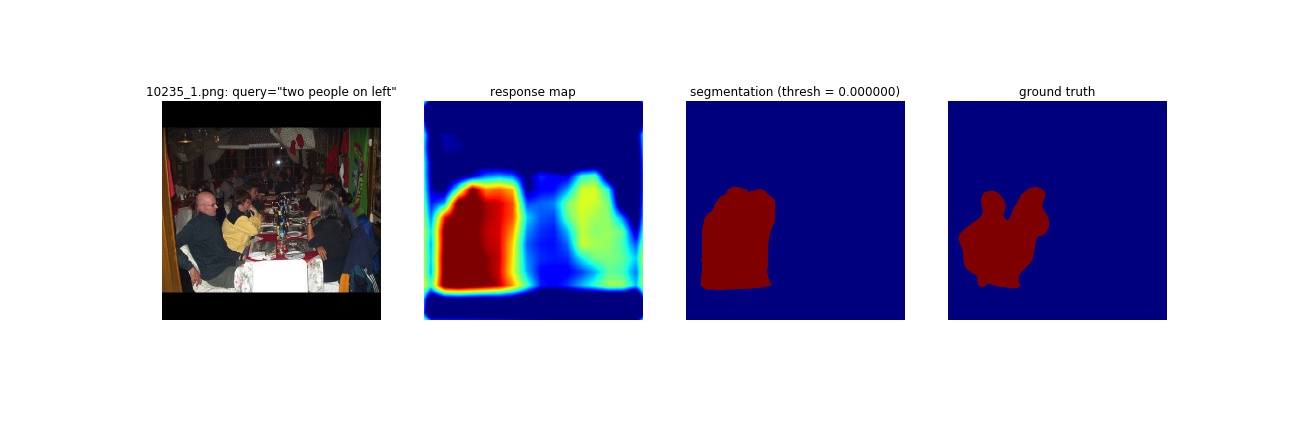} \\
\end{figure}

\begin{figure}[t]
\centering
\begin{tabularx}{0.95\linewidth}{YYYY}
input image & response map & our prediction & ground-truth \\ \hline
\end{tabularx} \\
\small{query expression=\refexp{stairs}} \\
\includegraphics[trim = 56mm 38mm 45mm 35mm, clip=true,width=0.90\textwidth,height=\textheight,keepaspectratio]{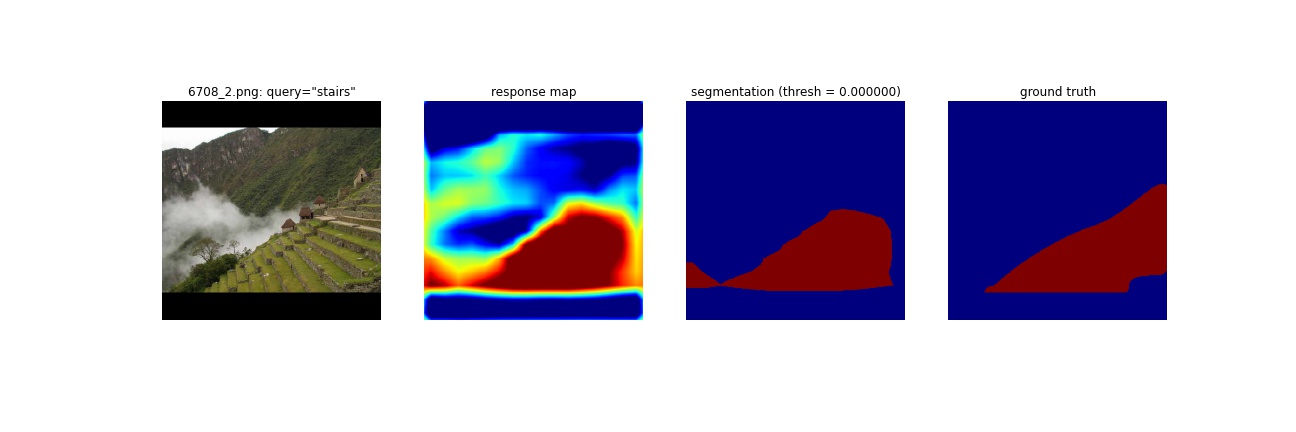} \\
\small{query expression=\refexp{brown brick walkway}} \\
\includegraphics[trim = 56mm 38mm 45mm 35mm, clip=true,width=0.90\textwidth,height=\textheight,keepaspectratio]{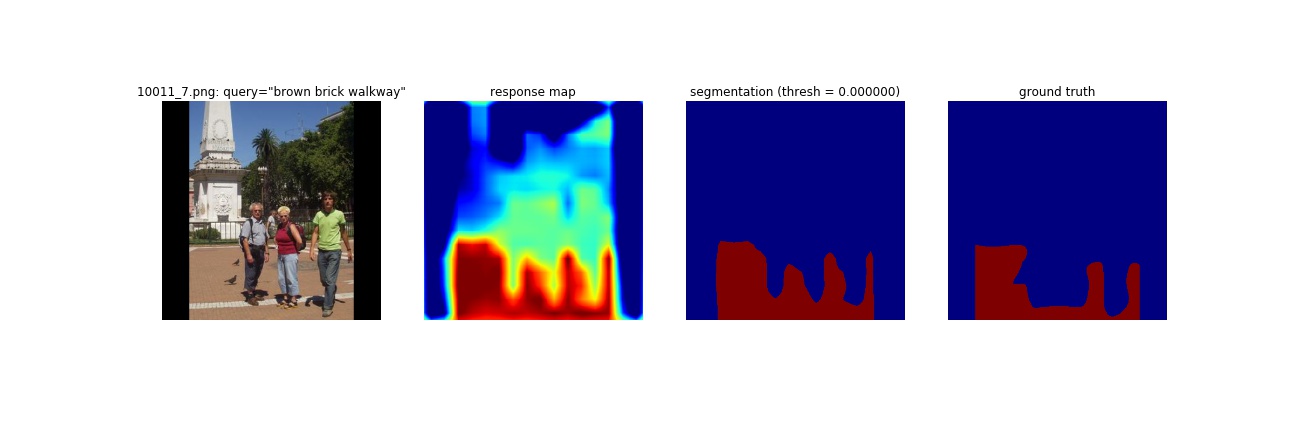} \\
\small{query expression=\refexp{court closest to us}} \\
\includegraphics[trim = 56mm 38mm 45mm 35mm, clip=true,width=0.90\textwidth,height=\textheight,keepaspectratio]{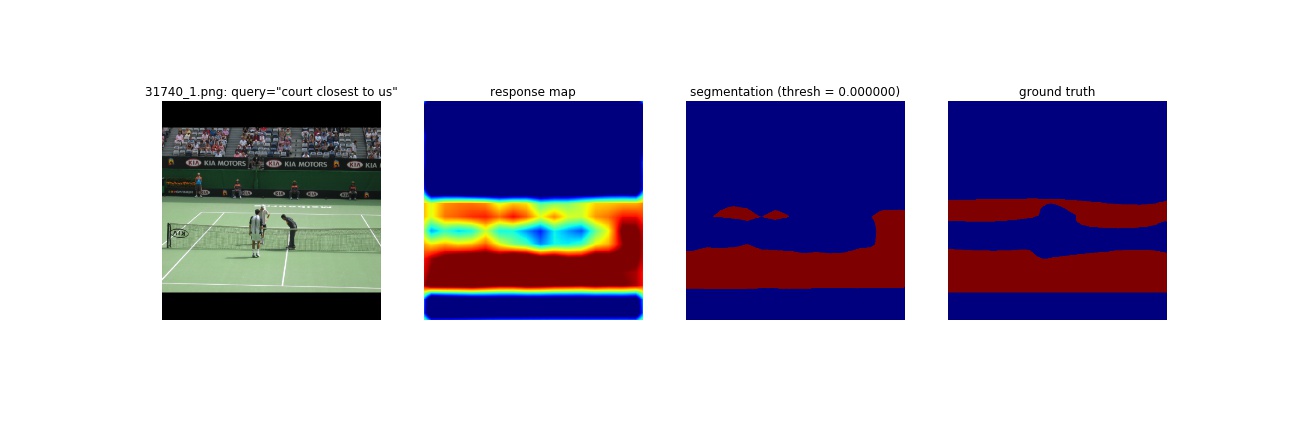} \\
\small{query expression=\refexp{trees background}} \\
\includegraphics[trim = 56mm 38mm 45mm 35mm, clip=true,width=0.90\textwidth,height=\textheight,keepaspectratio]{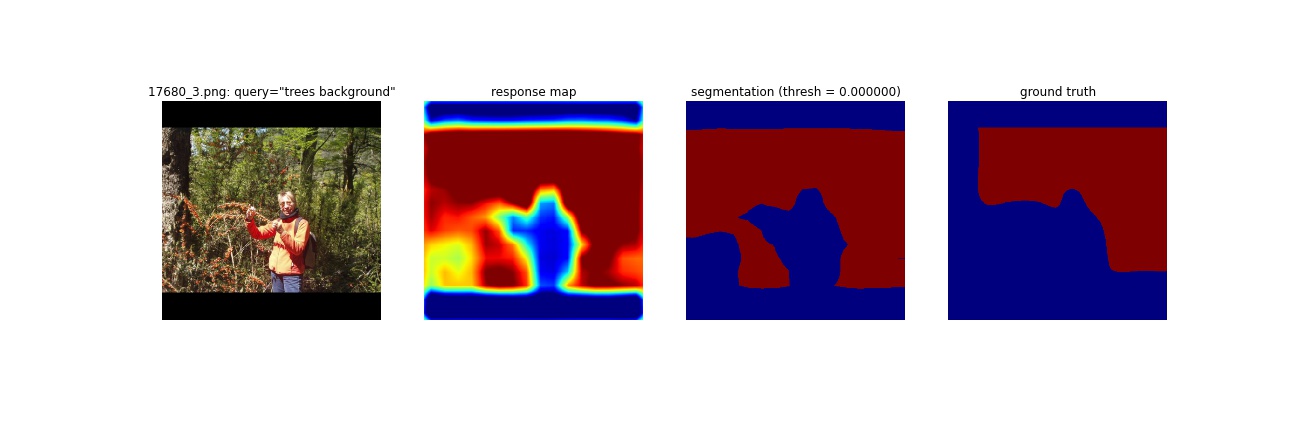} \\
\small{query expression=\refexp{the ruins}} \\
\includegraphics[trim = 56mm 38mm 45mm 35mm, clip=true,width=0.90\textwidth,height=\textheight,keepaspectratio]{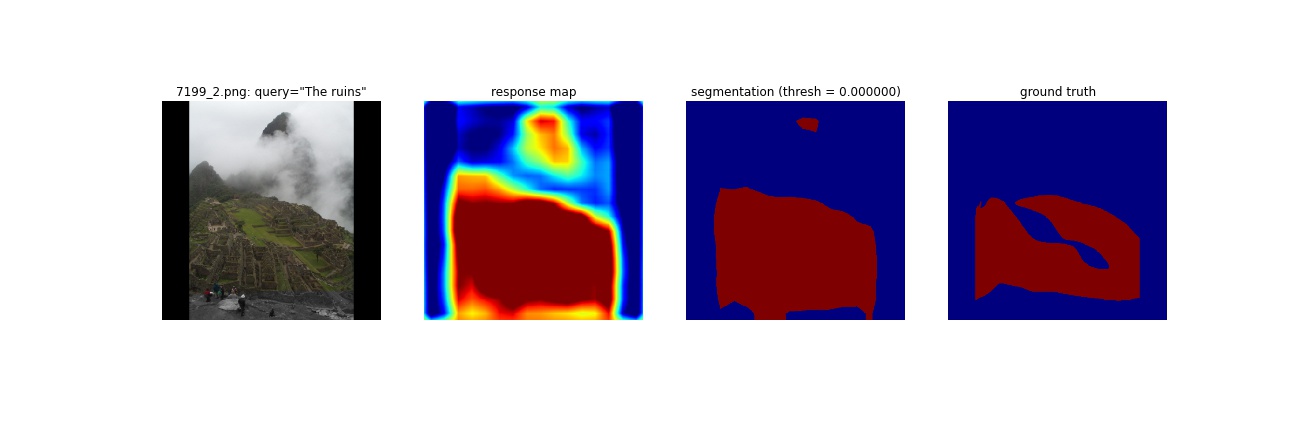} \\
\small{query expression=\refexp{bottom steps}} \\
\includegraphics[trim = 56mm 38mm 45mm 35mm, clip=true,width=0.90\textwidth,height=\textheight,keepaspectratio]{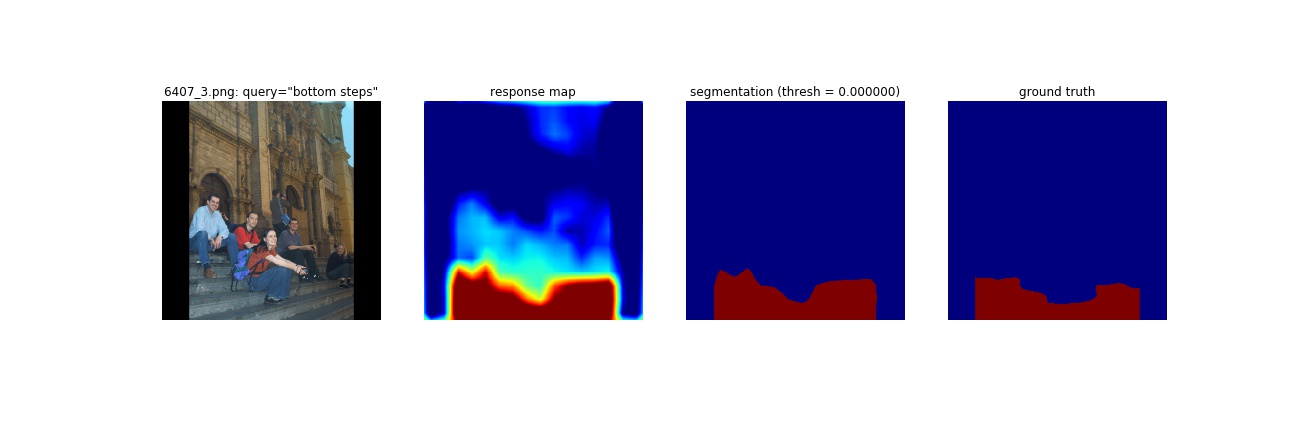} \\
\end{figure}

\end{document}

%% file: defines.tex
 \makeatletter
 \DeclareRobustCommand\onedot{\futurelet\@let@token\@onedot}
 \def\@onedot{\ifx\@let@token.\else.\null\fi\xspace}
 \def\eg{e.g\onedot} \def\Eg{E.g\onedot}
 \def\ie{i.e\onedot} \def\Ie{I.e\onedot}
 \def\cf{cf\onedot} \def\Cf{Cf\onedot}
 \def\etc{etc\onedot} \def\vs{vs\onedot}
 \def\wrt{w.r.t\onedot} \def\dof{d.o.f\onedot}
 \def\etal{\textit{et~al\onedot}} \def\iid{i.i.d\onedot}
 \def\Fig{Fig\onedot} \def\Eqn{Eqn\onedot} \def\Sec{Sec\onedot}
 \def\vs{vs\onedot}
 \makeatother

\DeclareRobustCommand{\figref}[1]{Figure~\ref{#1}}
\DeclareRobustCommand{\figsref}[1]{Figures~\ref{#1}}

\DeclareRobustCommand{\Figref}[1]{Figure~\ref{#1}}
\DeclareRobustCommand{\Figsref}[1]{Figures~\ref{#1}}

\DeclareRobustCommand{\Secref}[1]{Section~\ref{#1}}
\DeclareRobustCommand{\secref}[1]{Section~\ref{#1}}

\DeclareRobustCommand{\Secsref}[1]{Sections~\ref{#1}}
\DeclareRobustCommand{\secsref}[1]{Sections~\ref{#1}}

\DeclareRobustCommand{\Tableref}[1]{Table~\ref{#1}}
\DeclareRobustCommand{\tableref}[1]{Table~\ref{#1}}

\DeclareRobustCommand{\Tablesref}[1]{Tables~\ref{#1}}
\DeclareRobustCommand{\tablesref}[1]{Tables~\ref{#1}}

\DeclareRobustCommand{\eqnref}[1]{Equation~(\ref{#1})}
\DeclareRobustCommand{\Eqnref}[1]{Equation~(\ref{#1})}

\DeclareRobustCommand{\eqnsref}[1]{Equations~(\ref{#1})}
\DeclareRobustCommand{\Eqnsref}[1]{Equations~(\ref{#1})}

\DeclareRobustCommand{\chapref}[1]{Chapter~\ref{#1}}
\DeclareRobustCommand{\Chapref}[1]{Chapter~\ref{#1}}

\DeclareRobustCommand{\chapsref}[1]{Chapters~\ref{#1}}
\DeclareRobustCommand{\Chapsref}[1]{Chapters~\ref{#1}}

\hyphenation{po-si-tive}

%% file: macros.tex
\newcommand{\scream}[1]{\textbf{*** #1! ***}}
\newcommand{\fixme}[1]{\textcolor{red}{\textbf{FiXme}#1}\xspace}
\newcommand{\hobs}{\textrm{h}_\textrm{obs}}
\newcommand{\cpad}[1]{@{\hspace{#1mm}}}
\newcommand{\alg}[1]{\textsc{#1}}

\newcommand{\fnrot}[2]{\scriptsize\rotatebox{90}{\begin{minipage}{#1}\flushleft #2\end{minipage}}}
\newcommand{\chmrk}{{\centering\ding{51}}}
\newcommand{\eqn}[1]{\begin{eqnarray}\vspace{-1mm}#1\vspace{-1mm}\end{eqnarray}}
\newcommand{\eqns}[1]{\begin{eqnarray*}\vspace{-1mm}#1\vspace{-1mm}\end{eqnarray*}}

\newcommand{\myparagraph}[1]{\vspace{-0.25cm} \paragraph{\textbf{\emph{#1}}}}

\newcommand{\ronghang}[1]{\textcolor{red}{Ronghang: #1}}
\newcommand{\marcus}[1]{\textcolor{green}{Marcus: #1}}
\newcommand{\trevor}[1]{\textcolor{blue}{Trevor: #1}}
\newcommand{\invisible}[1]{}

\newcommand{\figvspace}{\vspace{-.5cm}}
\newcommand{\secvspace}{\vspace{-.2cm}}
\newcommand{\subsecvspace}{\vspace{-.2cm}}

\graphicspath{{./fig/}{./fig/plots/}}